\documentclass[runningheads]{llncs}
\usepackage{graphicx}
\usepackage[pagebackref=true,breaklinks=true,letterpaper=true,colorlinks,bookmarks=false]{hyperref}

\usepackage{times}
\usepackage{epsfig}
\usepackage{graphicx}
\usepackage[utf8]{inputenc} 
\usepackage[belowskip=5pt, aboveskip=0pt]{caption}
\usepackage{subcaption}
\usepackage{amsmath}
\usepackage{amssymb}

\usepackage{enumerate}
\usepackage{longtable}
\usepackage{booktabs}
\usepackage{pifont}
\usepackage{float}
\usepackage{psfrag}
\usepackage[percent]{overpic}

\newcommand{\xmark}{\ding{55}}%
\newcommand{\cmark}{\ding{51}}%
\usepackage[obeyFinal]{easy-todo}

\begin{document}
\def\MakeUppercaseUnsupportedInPdfStrings{\scshape}

\title{Multi-view Monocular Depth and Uncertainty Prediction with Deep SfM in Dynamic Environments\thanks{Supported by Robert Bosch GmbH.}}
\titlerunning{Learned Monocular Depth and Uncertainty from Dynamic Videos}
%
\author{Christian Homeyer \inst{1,2}\orcidID{0000-0002-0953-5162} \and
Oliver Lange \inst{1}\orcidID{0000-0001-7461-6869} \and
Christoph Schnörr \inst{2}\orcidID{0000-0002-8999-2338}}
\authorrunning{C. Homeyer et al.}
%
\institute{
	Robert Bosch GmbH, Hildesheim, Germany \email{\{christian.homeyer\}@bosch.com}\and
	IPA Group, Heidelberg University, Germany
}
\maketitle              
\begin{abstract}
 	3D reconstruction of depth and motion from monocular video in dynamic environments is a highly ill-posed problem due to scale ambiguities when projecting to the 2D image domain. 
 	In this work, we investigate the performance of the current State-of-the-Art (SotA) deep multi-view systems in such environments. 
 	We find that current supervised methods work surprisingly well despite not modelling individual object motions, but make systematic errors due to a lack of dense ground truth data. 
 	To detect such errors during usage, we extend the cost volume based Deep Video to Depth (DeepV2D) framework \cite{teed2018deepv2d} with a learned uncertainty. 
 	Our Deep Video to certain Depth (DeepV2cD) model allows i) to perform en par or better with current SotA and ii) achieve a better uncertainty measure than the naive Shannon entropy. 
 	Our experiments show that a simple filter strategy based on the uncertainty can significantly reduce systematic errors. This results in cleaner reconstructions both on static and dynamic parts of the scene.
\end{abstract}

%
%
%
\section{Introduction}
Reconstruction of scene geometry and camera motion is an important task for autonomous driving and related downstream tasks, e.g. collision avoidance and path planning. In particular, reconstruction from \textit{monocular} videos has become an important research direction, due to the possibility of low-cost realizations in concrete autonomous systems.

\begin{figure}[h]
	\centering
	 \captionsetup{skip=10pt}
	 \setlength{\textfloatsep}{0.0\baselineskip} 
	\includegraphics[height=6cm]{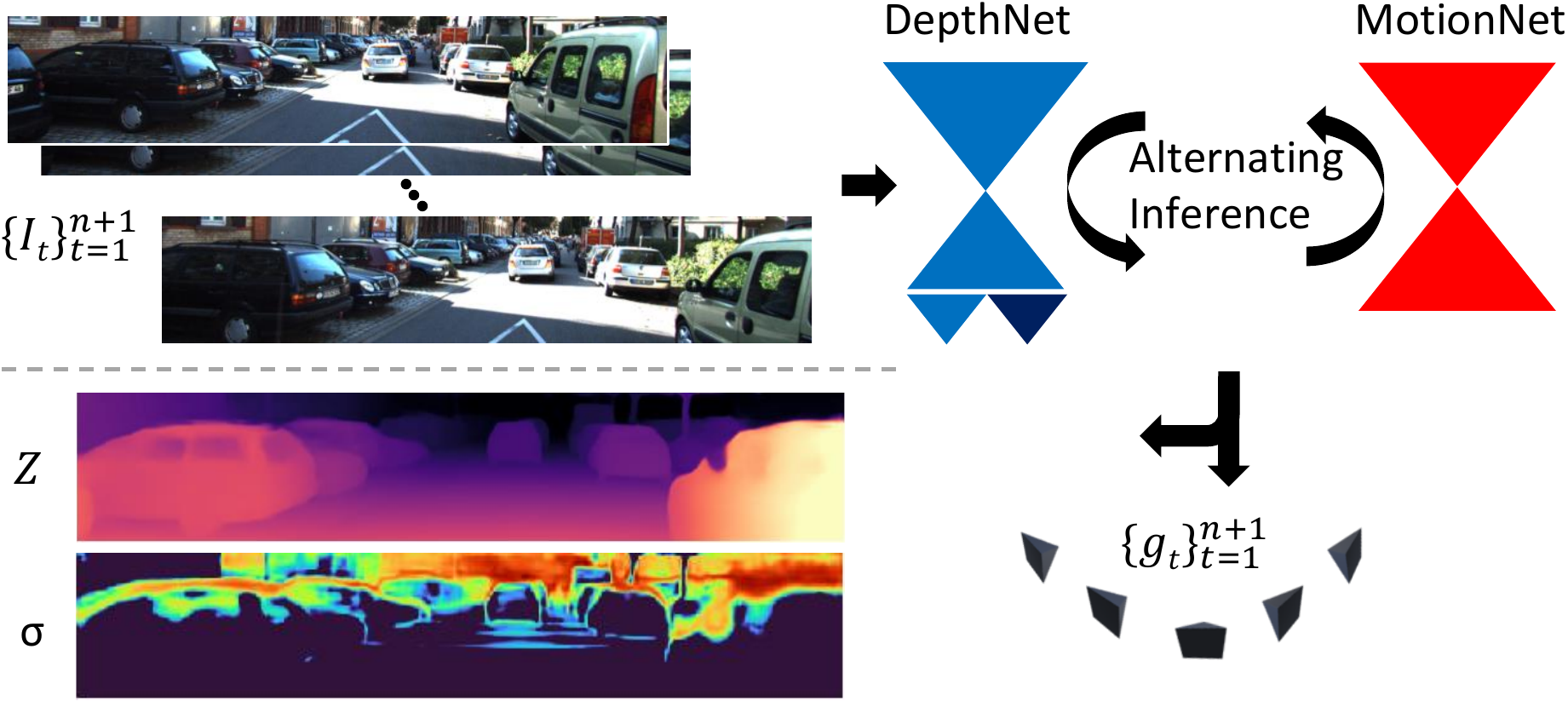}
	\caption{
		Based on an image sequence, we predict camera pose graph, depth and uncertainty. Warm colors indicate high uncertainty. 
		The uncertainty prediction helps to identify falsely reconstructed regions, which typically occur due to missing lidar supervision.
	}
	\vspace{-10pt}
	\label{fig:eyecatcher}
\end{figure}

\textit{Traditional} non-learning algorithms \cite{wu2011visualsfm} either fail to resolve the inherent ambiguities resulting from moving objects or do not scale well for dense predictions \cite{avidan1999trajectory}. 
\textit{Deep learning} based systems can resolve these shortcomings, but their accuracy is dependent on architecture and learning strategy. Depth reconstruction from a single image \cite{eigen2014depth,liu2015learning,li2015depth,laina2016deeper,fu2018deep} is an even more challenging problem, as no motion cues can act as input. \textit{Single-view} networks may not generalize well from one dataset to another \cite{liu2015learning,dijk2019neural} or need to be trained on massive datasets \cite{ranftl2019towards,ranftl2021vision}. Using a true \textit{multi-view} learning approach \cite{ummenhofer2017demon,tang2018ba,teed2018deepv2d,watson2021temporal,gu2021dro} turned out to be favorable in terms of performance. An \textit{unsupervised} learning strategy can leverage large-scale datasets without gathering expensive depth groundtruth. 
While unsupervised learning in this area has seen impressive progress \cite{garg2016unsupervised,godard2017unsupervised,godard2019digging,guizilini20203d,watson2021temporal,gu2021dro}, accurate depth on dynamic objects cannot be learned when objects move colinear to the camera, since the photometric loss cannot resolve the underlying ambiguity \cite{yuan2007detecting}. 
\textit{Supervised} training can leverage a signal for these cases, which is why we base our work on a supervised multi-view framework. 

Recent developments have shown, that modeling scene geometry explicitly inside the architecture \cite{teed2018deepv2d,tang2018ba,gu2021dro} leads to better reconstruction results than loosely coupling neural networks with a common training loss. Inside this paradigm, learned features are aligned temporally and spatially based on scene geometry and camera motion. 
Our evaluation in Section \ref{seq:experiments} indicates, that depth prediction works surprisingly well for dynamic objects, even though a static scene is assumed during this explicit modelling. However, we demonstrate quantitatively and qualitatively that reconstructions of dynamic objects are systematically worse than the static scene on real world autonomous driving data, e.g. the KITTI dataset. 
Our findings show, that the root cause is mostly a lack of supervision from the lidar ground truth. Lidar density is significantly lower on moving objects than on the rest of the scene and is missing completely near depth discontinuities or on transparent surfaces, e.g. windows.

\paragraph{}
We extend the supervised DeepV2D framework \cite{teed2018deepv2d} by learning an additional uncertainty and show how this can identify gross outliers in these regions. An overview of our DeepV2cD framework can be seen in Figure \ref{fig:eyecatcher}: based on a sequence of images we compute a) a camera pose graph b) the key frame depth and c) an uncertainty estimate. 
Because the depth network utilizes a cost volume based architecture, this extension comes with no additional overhead. 
In fact, any cost volume based approach maintains a probability volume, such that an uncertainty measure can be readily computed \cite{liu2019neural,zhang2020adaptive}.
However, our experiments indicate, that by adding a separate uncertainty head to our network and learning a measure, we achieve superior performance compared to the Shannon entropy as done in related work \cite{yang2019inferring,liu2019neural,laidlow2020towards}. Training this uncertainty head requires an additional loss term, which we adapt from \cite{zhang2020adaptive}. Our training strategy results in slightly better final accuracy and has the benefit of an uncertainty. 

An alternative approach to dealing with missing groundtruth in real data, is the use of synthetic data. Advances in datasets \cite{mayer2016large,dosovitskiy2017carla,mayer2018makes,cabon2020virtual} allow training with a dense and accurate ground truth before finetuning unsupervised on sparse real data \cite{yoon2020novel}. As a final contribution we show, that the Deepv2D networks generalize well to the real world even when only trained on virtual data. The domain gap can be further reduced by finetuning with a semi-supervised loss on the real data without the need for expensive depth ground truth. We make a comparison of current unsupervised, weakly supervised and supervised SotA on the KITTI and Cityscapes dataset.

\section{Related work}
\label{sec:related-work}
Reconstruction from monocular image sequences has a large body of literature. We refer interested readers to an in-depth survey with focus on dynamic scenes \cite{saputra2018visual}, that is beyond the scope of this paper. 
We distinguish between \textit{traditional} methods and \textit{learning} based systems. Learned methods can be distinguished into loosely coupled individual networks with \textit{single-view} depth prediction and true \textit{multi-view} systems. Finally, we give an overview of methods for \textit{uncertainty} prediction for this problem. 

\paragraph{Traditional.}
Early work in SfM worked only on a small collection of images \cite{longuet1981computer}. Traditionally, the problem is solved by finding correspondences with hand-crafted features, 
then solving for motion and structure. Today, several mature full pipelines exist \cite{wu2011visualsfm,snavely2008scene,schonberger2016structure}, that improved considerably w.r.t scalability, robustness and accuracy compared to early systems. The reconstruction  of monocular images is only possible to a common relative scale, due to a missing absolute scale. 
Traditional SfM relies on a static scene assumption. By filtering out dynamic points as outliers, they do not perform a full scene reconstruction. Several non-learning approaches target these shortcomings: Avidan et al. \cite{avidan2000trajectory,shashua1999trajectory,avidan1999trajectory} coined the term trajectory triangulation by extending the epipolar constraint to the general case of a moving point with a specific trajectory. Park et al. \cite{akhter2008nonrigid,park20103d,park20153d} show how it is possible to decompose motions into a linear combination of basis trajectories using the Discrete Cosine Transform (DCT). However, these approaches only perform sparse reconstruction and require knowing the dynamic/static assignment to be known in advance. Another line of work relies on the piecewise-planar assumption and dense reconstruction based on optical flow \cite{ranftl2016dense,kumar2017monocular,kumar2019superpixel}. Still, especially colinearly moving objects cannot be distinguished from observed background and result in wrong reconstructions \cite{yuan2007detecting}.

\paragraph{Learning geometry and motion.}
Early work on learning reconstructions used e.g. Markov Random Fields (MRF's) \cite{saxena2008make3d,liu2010single} or non-linear classifiers \cite{ladicky2014pulling}. 
With the advances of deep neural networks, the single-view reconstruction problem can be effectively approached with Convolutional Neural Networks (CNN's) \cite{eigen2014depth,liu2015learning,li2015depth,laina2016deeper,fu2018deep,kendall2018multi}. 
Camera poses can also be inferred by deep neural networks. \cite{kendall2015posenet,kendall2016modelling} learn camera poses from single images, while \cite{wang2017deepvo,yin2018geonet,chen2019self} learn from image pairs. By learning the poses from whole sequences \cite{clark2017vinet}, \cite{wang2017deepvo}, \cite{yang2020d3vo}, a more accurate camera trajectory can be estimated. An important trend is the development of self-supervised learning objectives for learning both geometry and motion \cite{garg2016unsupervised,godard2017unsupervised,zhou2017unsupervised,yang2018every,godard2019digging,johnston2020self,guizilini20203d}, which alleviates the need of expensive 3D ground truth data. In this context, dynamic objects were adressed with semantics \cite{casser2019unsupervised,lee2019instance,klingner2020self} or general motion parametrizations and object masks \cite{vijayanarasimhan2017sfm,li2020unsupervised,ranjan2019competitive}. However, self-supervised training cannot fully resolve the inherent ambiguities in dynamic scenes. Furthermore, they rely on single-view reconstruction networks, which may not generalize well from one dataset to another \cite{liu2015learning,dijk2019neural} or create the need for massive datasets \cite{ranftl2019towards}. Ranftl et al. \cite{ranftl2021vision} apply the transformer architecture to supervised single-view depth prediction and achieve SotA results within this setting, but the accuracy is still behind multi-view approaches on our target datasets. Another line of work focuses on 
improving temporal depth consistency in dynamic scenes by finetuning existing single-view networks \cite{kopf2021robust,zhang2021consistent}. While they improve the accuracy considerably, they rely on a complicated processing pipeline and use an offline camera pose estimation. Other works learn geometry from multi-view information based on cost volumes \cite{yao2018mvsnet,liu2019neural,duzceker2021deepvideomvs,sun2021neuralrecon,bovzivc2021transformerfusion}, but assume known camera poses and no object motion. Wimbauer et al. \cite{wimbauer2020monorec} leverage a cost volume in a semi-supervised approach. They learn a motion mask for dynamic objects and use stereo supervision in these areas. 

\paragraph{Combinations with traditional SfM.}
One of the first deep SfM systems was published by \cite{ummenhofer2017demon}. Since then, a series of frameworks combine multi-view image information for inferring camera motion and scene geometry \cite{zhou2018deeptam,clark2018learning,tang2018ba,teed2018deepv2d,gu2021dro,watson2021temporal}. While most works rely on generic network architectures, few combine learning with a 
traditional geometric optimization \cite{teed2018deepv2d,tang2018ba,clark2018learning}. We base our model on DeepV2D \cite{teed2018deepv2d},  which couples supervised training of depth based on a cost volume architecture with a geometric pose graph optimization. These choices allow the networks to exhibit a strong cross-dataset generalization, which can be seen from the results in \cite{teed2018deepv2d} and our experiments. 
Recent work Manydepth focuses on training unsupervised multi-view cost volumes \cite{watson2021temporal}, but unsupervised training results in larger errors on moving objects due to the beforementioned problems. Recent work DRO \cite{gu2021dro} has focused on creating a fully recurrent architecture both for structure and motion estimation. 
While DRO collapses the cost volume in order to achieve a lightweight architecture, we exploit the uncertainty information that is maintained inside the volume. We focus on depth prediction, but our approach can also be leveraged in downstream tasks, e.g. scene flow estimation \cite{brickwedde2019mono} or mapping/fusion \cite{liu2019neural,zhou2018deeptam,laidlow2020towards}.

\paragraph{Uncertainty estimation.}
Uncertainty can be distinguished into \textit{aleatoric} and  \textit{epistemic} uncertainty \cite{kendall2017uncertainties}. While epistemic uncertainty is the uncertainty over model parameters, aleatoric uncertainty is over the model outputs. 
Several uncertainty estimation strategies exist for regression problems: 1. Learn parameters of probability distribution as output \cite{kendall2017uncertainties,yang2019inferring,brickwedde2019mono}. 2. Maintain a full discrete probability volume \cite{kendall2017end,teed2018deepv2d,liu2019neural,zhang2020adaptive,laidlow2020towards} 3. Bayesian neural network with distributions over model parameters 4. Dropout variational inference \cite{kendall2017uncertainties}. Another related work is \cite{poggi2020uncertainty}, which investigates various uncertainty techniques for unsupervised single-image depth prediction. Our approach falls in the second category and learns an aleatoric uncertainty. The idea of using cost volumes for depth reconstruction originates from stereo disparity estimation \cite{kendall2017end,chang2018pyramid,zhang2020adaptive,zhang2019ga}. The uncertainty was not yet investigated and exploited with more recent deep SfM-frameworks in the multi-view setting. We realize, that using the Shannon entropy naively \cite{liu2019neural,yang2019inferring} within our framework does not result in optimal uncertainty estimates. We instead follow a similar strategy as \cite{zhang2020adaptive}, and introduce an uncertainty head into the network.

\begin{figure*}[h!]
	\centering
	\includegraphics[width=1.0\linewidth]{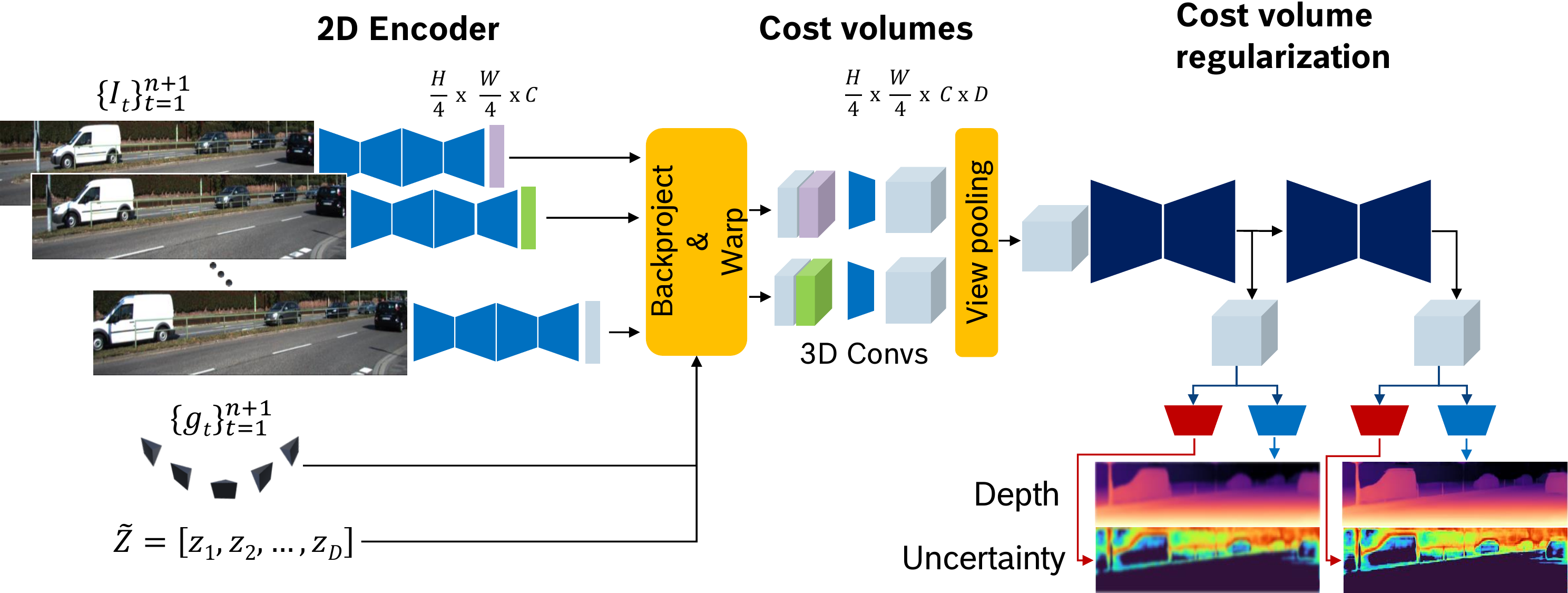}
	\caption{Depth network of DeepV2cD. Based on $ \{ I_t \}^{t=n+1}_{t=1} $ and current estimated camera poses$ \{ g_t \}^{t=n+1}_{t=1} $ the network builds a cost volume 
	with learned features for each frame. 3D hourglasses regularize the cost volume and output intermediate volumes. We train a separate uncertainty head to learn an 
	uncertainty $ \sigma $ for the depth estimate.}
	\label{fig:depth network}
\end{figure*}

\section{Approach}
\label{sec:approach}
\subsection{Problem Statement}
Given an ordered sequence of images $ \{ \textbf{I}_t \}^{2n+1}_{t=1} $ from a calibrated camera with intrinsics $ \textbf{K} $, compute the set of extrinsic camera parameters $ \{ \textbf{g}_t \}^{2n+1}_{t=1} $ and scene depth $ Z_{n+1} $. This can be considered as windowed bundle adjustment problem. For ease of notation, we use simply $ Z $ to denote the key frame depth.

\subsection{Notation}
\label{sec:notation}
Based on a pinhole camera model, projection and backprojection are defined as:
\begin{align}
\pi\left(\textbf{X} \right) &= 
\begin{bmatrix}
f_{x}\frac{X}{Z} + c_{x}, & f_{y}\frac{Y}{Z} + c_{y} \\
\end{bmatrix} \\
\pi^{-1}\left(\textbf{x},\; Z \right) &= 
\begin{bmatrix}
Z\frac{u-c_{x}}{f_{x}}, & Z\frac{v-c_{y}}{f_{y}}, & Z, & 1
\end{bmatrix}^{T}
\label{eq:projection}
\end{align}
with camera coordinates $ \textbf{x} = \left[ u,\; v,\; 1 \right] $, 3D coordinates $ \textbf{X} = \left[ X,\; Y,\; Z,\; 1 \right]^{T} $ and 
camera intrinsics $ \left[ f_{x},\; f_{y},\; c_{x},\; c_{y} \right] $. The camera motion is modelled using the rigid body motion $ \textbf{g}_{ij} \in SE_{3} $. Given two views $ i $ and $ j $, the relative coordinate transformation is defined as:
\begin{align}
\textbf{x}^{j} = \pi\left( \textbf{g}_{ij} \textbf{X}^{i} \right) &= \pi\left( \textbf{g}_{j}\textbf{g}^{-1}_{i} \pi^{-1}\left( \textbf{x}^{i},\; Z \right)  \right) \\
& = \Psi\left( \textbf{g}_{ij},\; \textbf{x}^{i},\; Z \right) ,\; \nonumber
\label{eq:view2view}
\end{align}
where $ \Psi $ is the reprojection operator that transforms a pixel coordinate from camera $ i $ into camera $ j $. Given an element of the Lie group $ \textbf{g} $, the logarithm map $ \xi = \log \textbf{g} \in se_{3} $ maps to the Lie algebra. Vice versa the exponential map is defined as $ \textbf{g} = e^{\xi} \in SE_{3} $.

\subsection{Networks}
\label{sec:networks}
This section introduces the DeepV2cD framework. After restating the base networks \cite{teed2018deepv2d}, 
we introduce our uncertainty head and the explored learning strategies.

\paragraph{Motion Network.}
We use the motion network from \cite{teed2018deepv2d}. We briefly summarize the architecture and idea in the following. 
The motion network predicts a set of camera poses $ \{ \textbf{g}_t \}^{2n+1}_{t=1} $ based on $ Z $ and $ \{ \textbf{I}_t \}^{2n+1}_{t=1} $. It is a recurrent network that instead of regressing a final pose set, outputs a sequence of $ l $ sets $ \left[  \{ \textbf{g} \}_{1},\; \{ \textbf{g} \}_{2},\; \dots,\; \{ \textbf{g} \}_{l} \right] $. The camera poses $ \{ \textbf{g}_t \}^{n+1}_{t=1} $ act as state variables, that determine the alignment of learned features and are updated based on an internal geometric pose graph optimization. 
Relative camera poses between frames are initialized based on a generic network \cite{zhou2017unsupervised}. A feature encoder produces dense features $ F^{i} \in \mathbb{R}^{\frac{H}{4} \times\frac{W}{4} \times 32} $ for each image $ I^{i} $. Based on the current state $ \{ \textbf{g}_{t,i} \}^{2n+1}_{t=1} $ and input scene geometry $ Z $ features are aligned in a canonical coordinate frame and concatenated. A 2D hourglass network \cite{newell2016stacked} estimates a pairwise residual optical flow $ \textbf{R} $ and confidence map $ \textbf{W} $, depending on the variable state across recurrences. We observe, that the motion network learns to ignore regions in the image, that violate the static scene assumption via the confidence $ \textbf{W} $. It only keeps regions with good optical flow for pose optimization. Given a state $ \{ \textbf{g}_{t} \}^{2n+1}_{t=1} $, $ Z $, $ \textbf{W} $ and $ \textbf{R} $, an update $ \xi $ can be computed via Gauss-Newton optimization \cite{teed2018deepv2d}.

\paragraph{Depth Network.}
We extend the depth architecture with an additional prediction head and restate the core principles in the following. An overview of the depth network can be seen in Figure \ref{fig:depth network}. The network design is conceptually similar to PSMNet \cite{chang2018pyramid}, but gives only two instead of three intermediate predictions due to memory constraints in our setting. This network predicts a dense depth map $ Z $ based on $ \{ \textbf{I}_t \}^{2n+1}_{t=1} $ and $ \{ \textbf{g}_t \}^{2n+1}_{t=1} $. The motion of dynamic objects is not given as input to the network. 

For each image $ I^{i} \in \mathbb{R}^{H\times W \times 3} $, a 2D hourglass encoder \cite{newell2016stacked} learns features $ F^{i} \in \mathbb{R}^{\frac{H}{4}\times \frac{W}{4}\times L} $ with feature dimension $ L=32 $. The 2D features for each time frame $ i $ are reprojected based on the static scene geometry and camera motion defined by $ (Z, \textbf{g}_{ij}) $ and aligned in a canonical coordinate frame $ j $, that is the keyframe. Given the correct depth $ Z^{*} $ and camera motions $ \textbf{g}_{ij}^{*} $, the features should be matched well as long as the scene is static. A cost volume $ \textbf{C} $ is constructed by backprojecting 2D features into the key frame over a discrete 1D depth interval with $ \tilde{Z} = \left[ z_{1}, z_{2}, \dots, z_{D} \right] $. The depth range in the scene is discretized into $ D-1 $ bins. When reprojecting features into this common reference frame based on $ \Psi $, they are bilinearly interpolated with the sampler $ \Phi\left(\cdot \right) $ \cite{jaderberg2015spatial}. For each point $ \textbf{x}^{i}_{k} $ in frame $ i $ and depth $ z_{d} \in \tilde{Z} $:
\begin{align}
C^{j}_{kd} &= \Phi \left[ \Psi_{ij} \left( \textbf{x}_{k}^{i},\; z_{d} \right) \right] \\
&= \Phi \left[ \pi\left( \textbf{g}_{j}\textbf{g}^{-1}_{i} \pi^{-1}\left( \textbf{x}_{k}^{i},\; z_{d} \right) \right) \right] \in \mathbb{R}^{\frac{H}{4}\times \frac{W}{4} \times D \times L} \;. \nonumber
\end{align}
We use a discretization of $ D=32 $ bins, ranging to 80m in practice due to memory constraints. The reprojection operator $ \Psi_{ij} \colon i \mapsto j $, is only dependent on camera motion and scene geometry. All non-keyframe features are concatenated with the backprojected keyframe features to form a final pairwise cost volume $ \textbf{C}^{ij} \in \mathbb{R}^{\frac{H}{4}\times \frac{H}{4} \times D \times 2L} $. Afterwards feature matching is learned with a series of 3D convolutions on the volume for each pair. The information across all pairs of images is globally aggregated in a pooling layer, so that a single cost volume $ \textbf{C} \in \mathbb{R}^{\frac{H}{4} \times \frac{W}{4} \times D \times L} $ remains. 
The cost volume is then processed by a series of 3D hourglass modules \cite{newell2016stacked}, which make up most of the networks parameters. This step is commonly referred to as cost volume regularization in the stereo literature \cite{kendall2017end,chang2018pyramid} and we believe it is crucial to achieve good performance in dynamic environments since up to this point the learned features have been matched solely on a static scene assumption. Each hourglass outputs an intermediate regularized cost volume that is run through a depth head to give a final depth estimate $ Z_{i} \in \mathbb{R}^{\frac{H}{4} \times \frac{W}{4} \times 1} $. 

The depth head produces a probability volume by a series of $ 1 \times 1 \times 1 $ convolutions and a softmax operator over the depth dimension. 
The prediction is estimated using the differentiable argmax function \cite{kendall2017end}, thus giving the expected depth of the probability distribution. All predictions are upsampled to final resolution $ H \times W $ with naive bilinear upsampling. The design is similar to the stereo network PSMNet \cite{chang2018pyramid}, but only gives two intermediate predictions instead of three in PSMNet due to memory constraints in the multi-view setting. 

Similar to related work \cite{yang2019inferring,laidlow2020towards,liu2019neural}, we can estimate an uncertainty based on the maintained probability volume. Given the probability distribution $ P $ over $ \tilde{Z} $, we can compute the Shannon entropy:
\begin{align}\label{eq:shannon}
	H = \sum_{d=1}^{D} P\left(z_{d}\right) \log P\left(z_{d}\right)
\end{align} as a measure of uncertainty \cite{yang2019inferring}. However, due to the limited discrete depth hypothesis space and the soft-argmax operation, the network does not necessarily learn to estimate uni-modal distributions since an infinite number of distributions can produce the same expected depth value. The entropy is not necessarily a good measure of uncertainty. Instead of taking the entropy naively, we learn a confidence $ f \in \left[ 0,\; 1\right]^{M \times N \times 1} $ with a separate network head similar to \cite{zhang2020adaptive}. 
This head performs a series of $ 3 \times 3 \times 3 $ convolutions on the regularized cost volume to check for hard-to-match pixels; we use 4 convolution layers. The uncertainty $ \sigma $ is then defined as the linearly scaled $ \sigma = s \cdot \left( 1 - f \right) + \epsilon $, which avoids numerical issues; we set $ s = 2 $ and $ \epsilon = 0.25$ empirically.
For training the uncertainty head we adopt the regularization loss from \cite{zhang2020adaptive} and define an unimodal groundtruth distribution:
\begin{align}
P^{*}\left( z_{d}\right) = softmax \left( - \frac{|z_{d} - Z^{*}|}{\sigma} \right)
\end{align} over all values in the depth hypothesis space $ z_{d} $ and with the estimated uncertainty $ \sigma $ from the uncertainty head. The sharpness of the peak is dependent on the uncertainty predicted by the network. 

\paragraph{Iterative inference by coupling motion and depth.}
As the two networks are functions partially dependent on the output of each other, it is alternated between them during inference. They converge to a final optimum after few iterations as shown in \cite{teed2018deepv2d}. 
At the start of the inference, the depth is initialized to a constant depth map for all of our experiments, thus requiring no warm start.

\begin{figure*}[h!]
	\centering
	\begin{overpic}[width=.238\linewidth, tics=10]
		{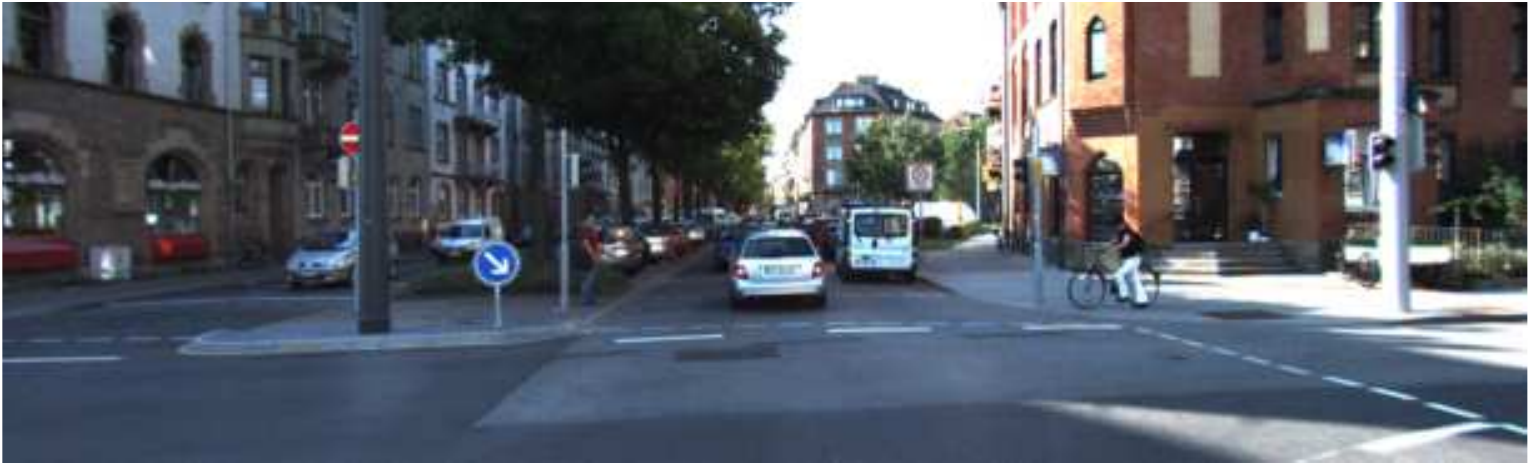}
		\put(-10, 4){\rotatebox{90}{\footnotesize{Data}}}
	\end{overpic}
	\begin{overpic}[width=.238\linewidth, tics=10]
		{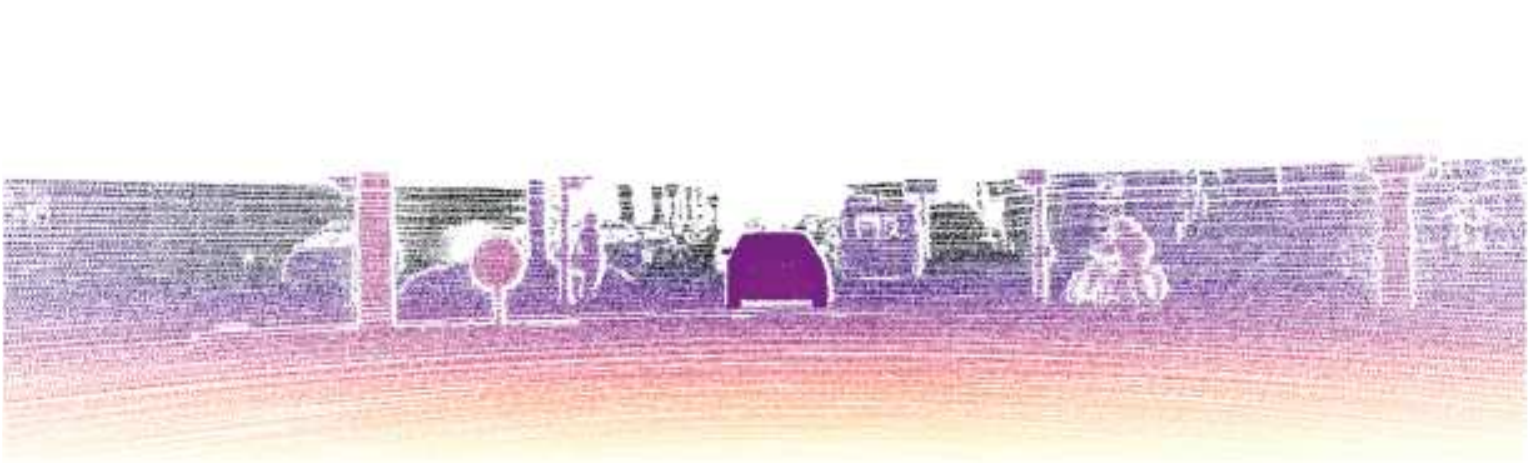}
		\put(10, 24){\footnotesize{Ground truth depth}}
	\end{overpic}
	\includegraphics[width=.238\linewidth]{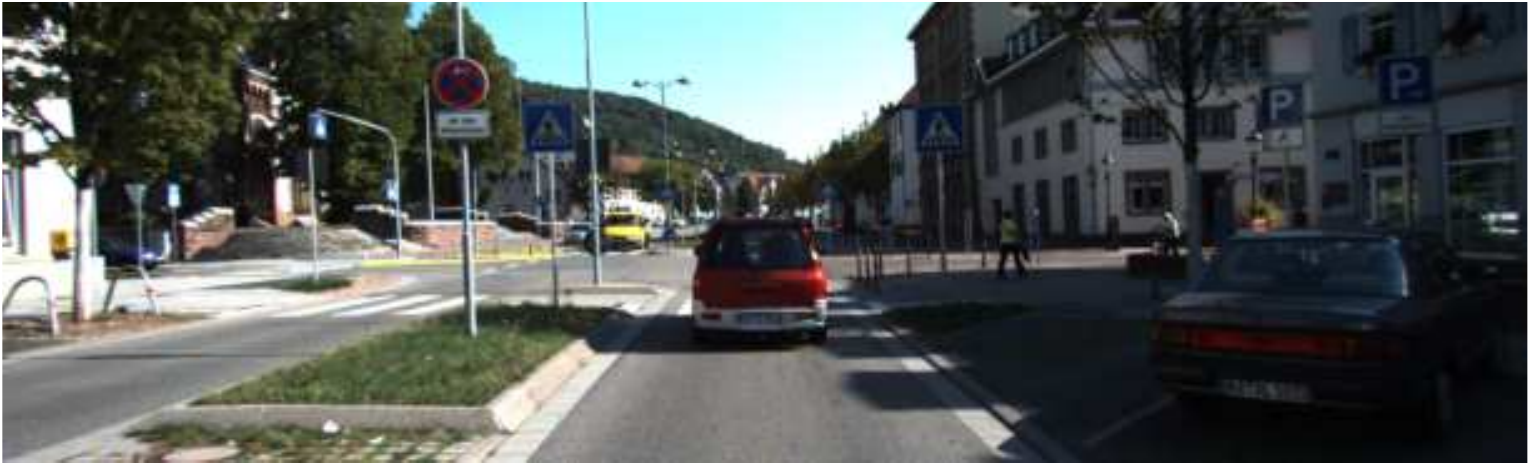}
	\begin{overpic}[width=.238\linewidth, tics=10]
		{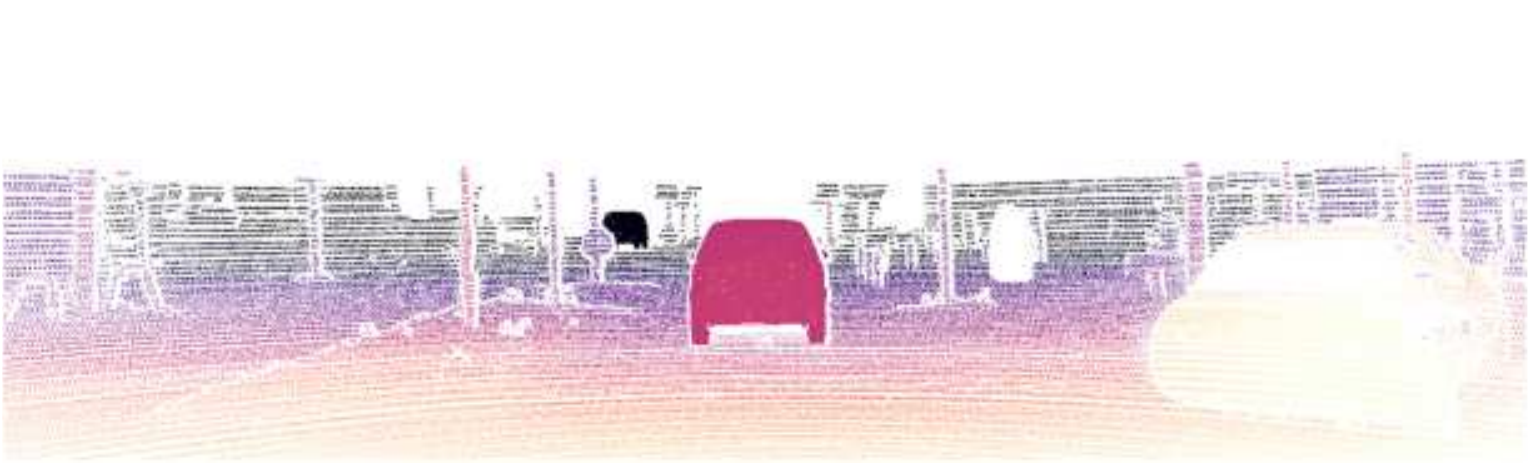}
		\put(10, 24){\footnotesize{Ground truth depth}}
	\end{overpic}
	\\[\smallskipamount]
	
	\begin{overpic}[width=.238\linewidth, tics=10]
		{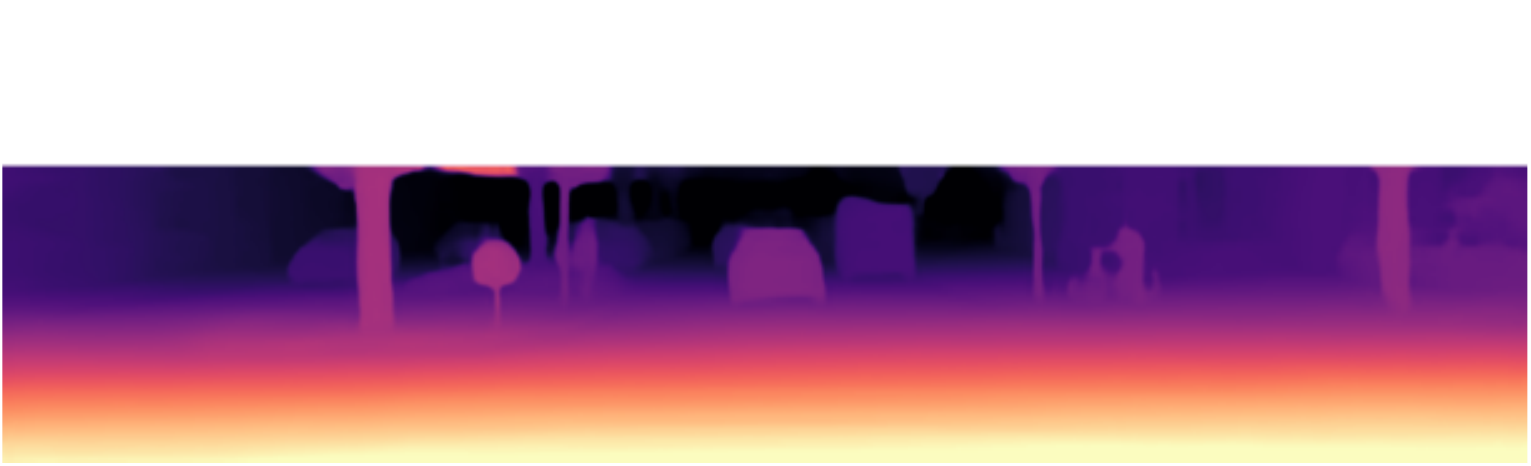}
		\put(20, 24){\footnotesize{Predicted depth}}
		\put(-10,7){\rotatebox{90}{\footnotesize{Ours}}}
	\end{overpic}
	\begin{overpic}[width=.238\linewidth, tics=10]
		{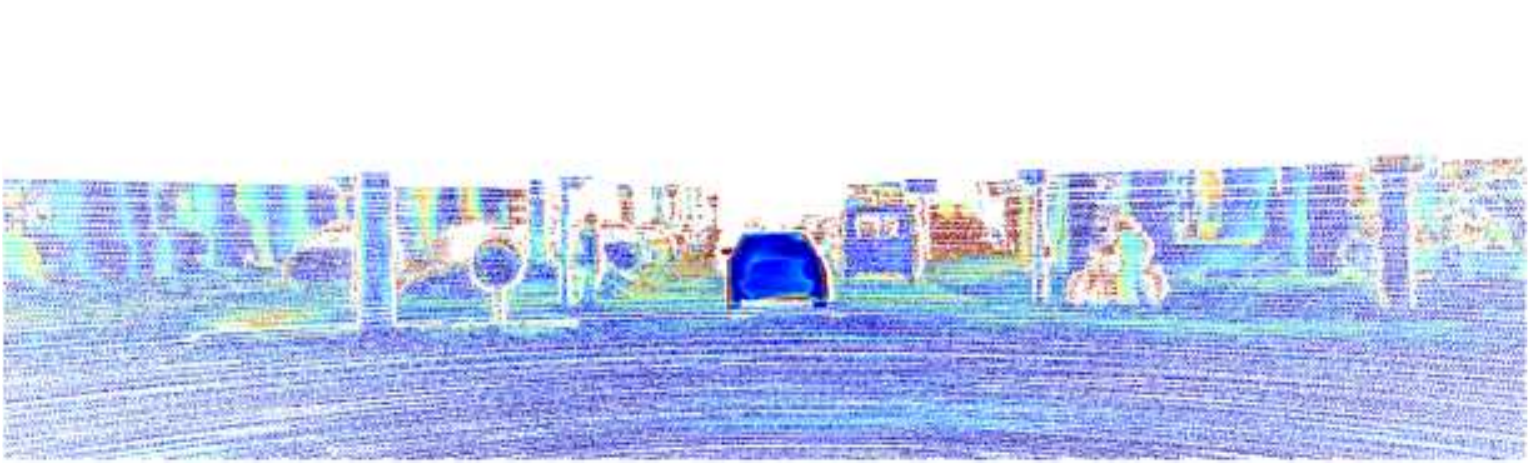}
		\put(20, 24){\footnotesize{Abs. rel. error}}
	\end{overpic}
	\begin{overpic}[width=.238\linewidth, tics=10]
		{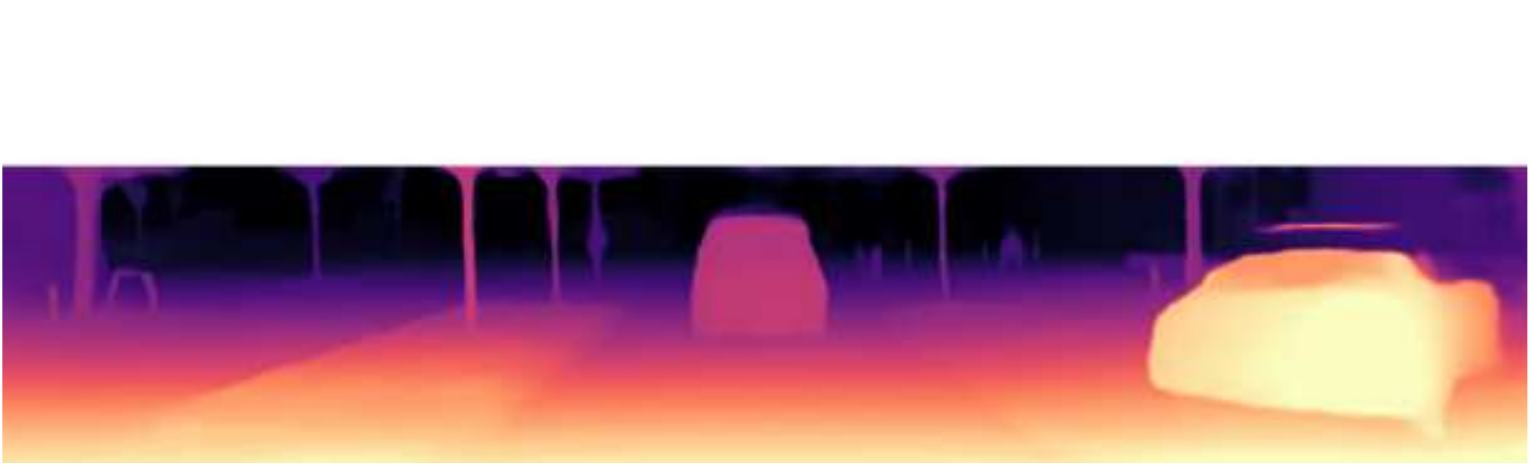}
		\put(20, 24){\footnotesize{Predicted depth}}
	\end{overpic}
	\begin{overpic}[width=.238\linewidth, tics=10]
		{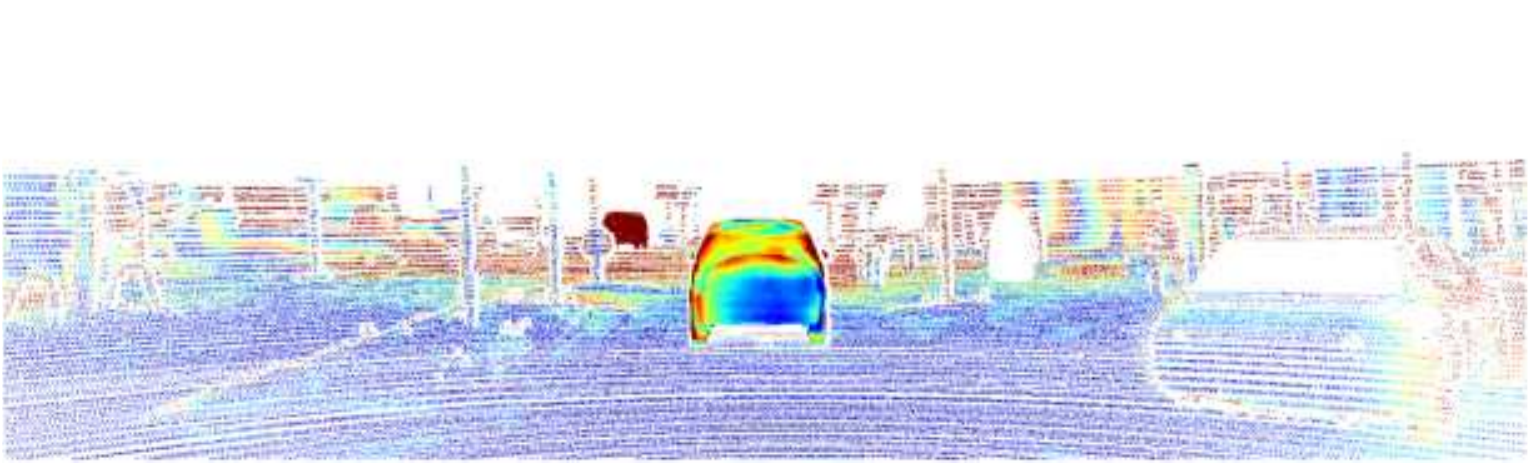}
		\put(28, 24){\footnotesize{Abs. rel. error}}
	\end{overpic}
	\\[\smallskipamount]
	
	\begin{overpic}[width=.238\linewidth, tics=10]
		{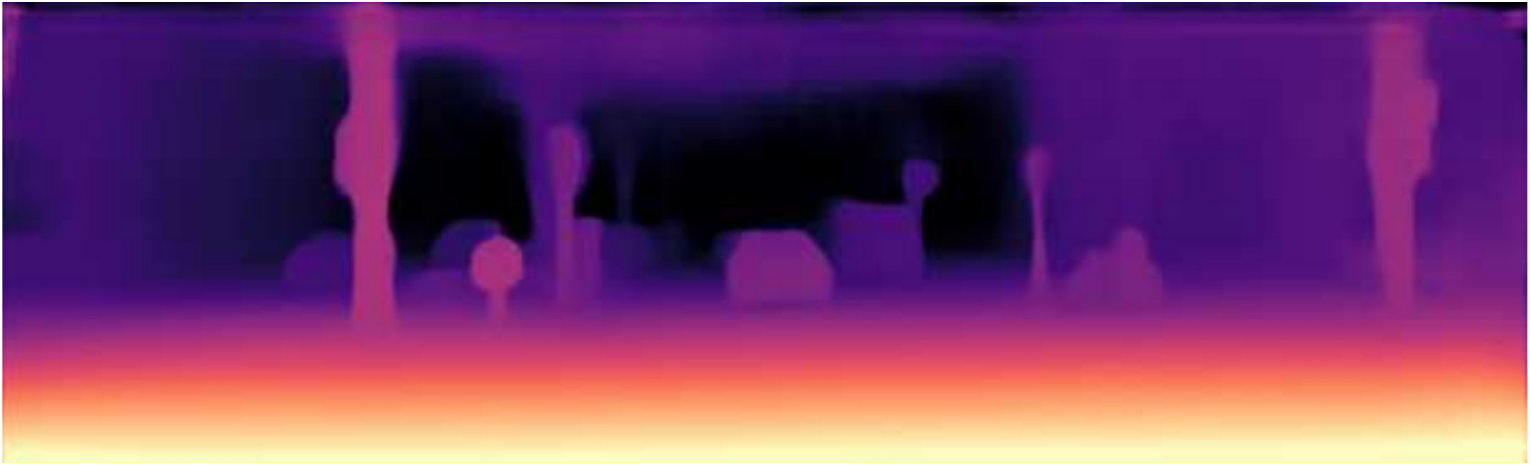}
		\put(-10,-6){\rotatebox{90}{\footnotesize{\cite{gu2021dro}spvd}}}
	\end{overpic}
	\includegraphics[width=.238\linewidth]{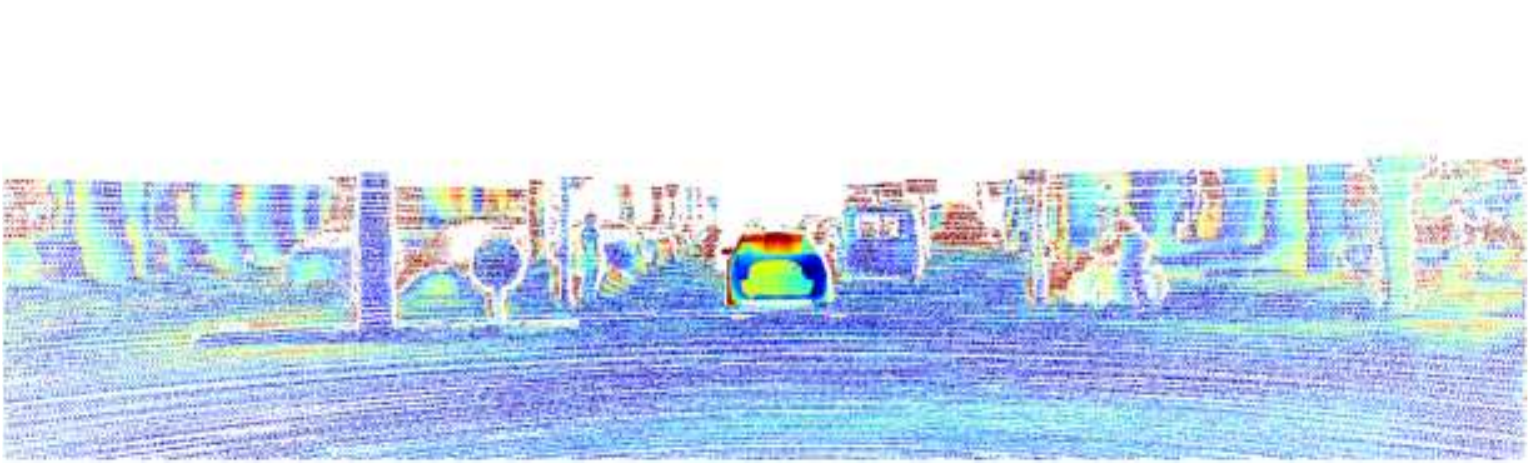}
	\includegraphics[width=.238\linewidth]{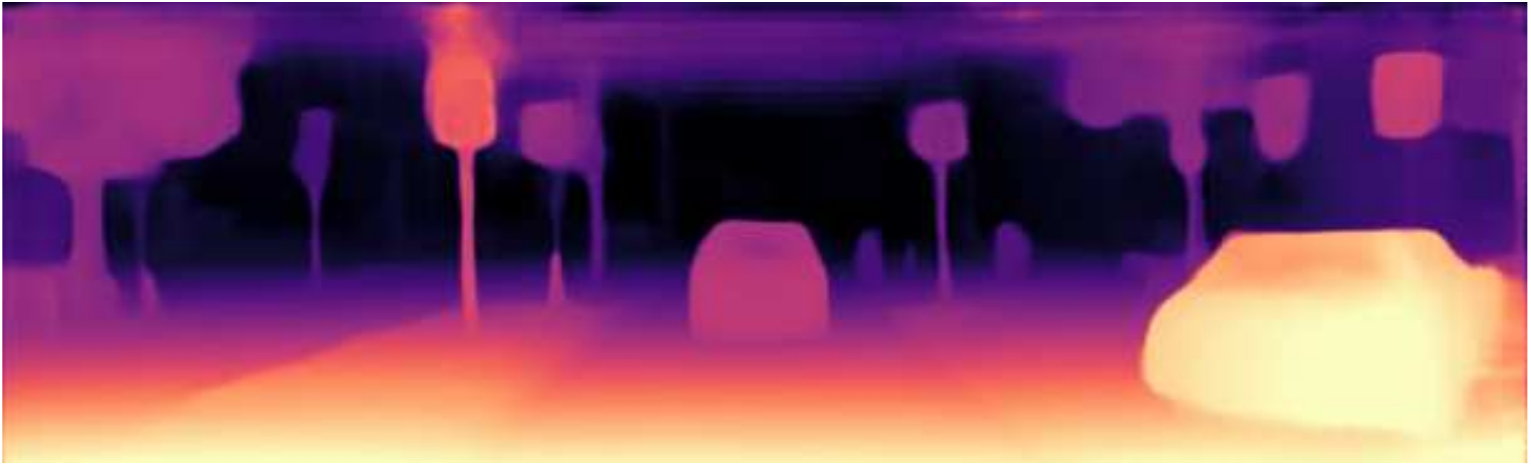}
	\includegraphics[width=.238\linewidth]{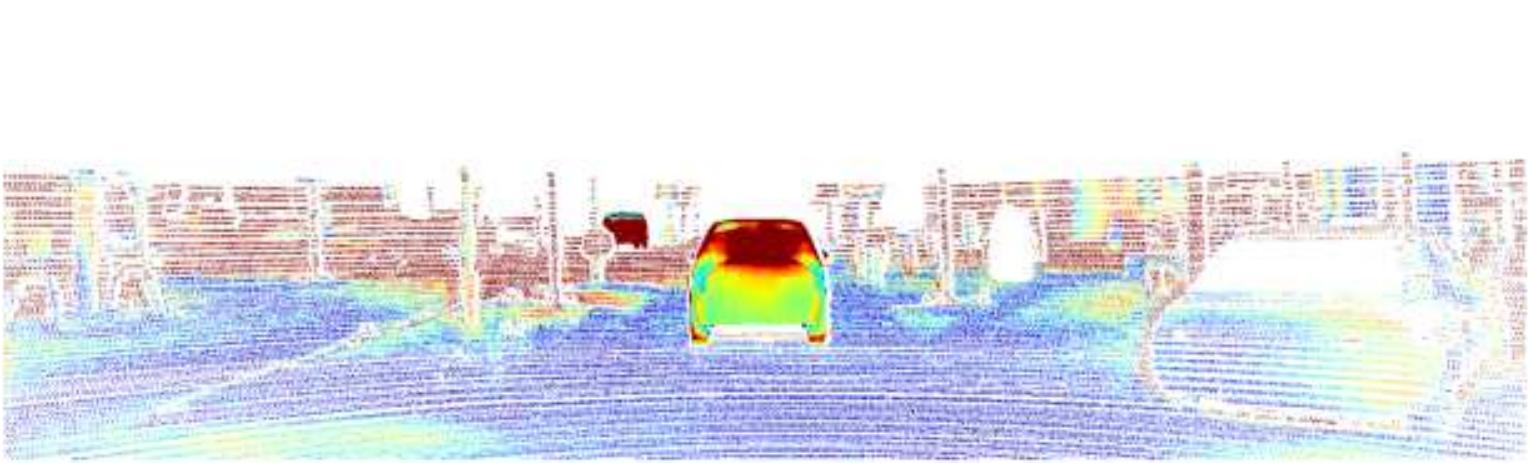}
	\\[\smallskipamount]
	
	\begin{overpic}[width=.238\linewidth, tics=10]
		{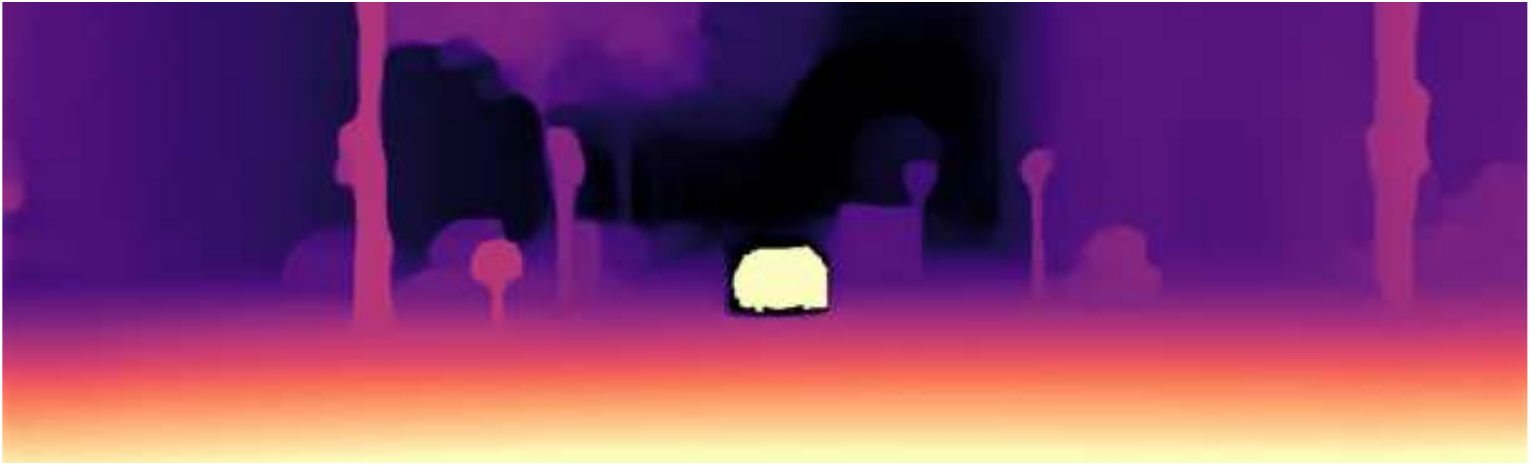}
		\put(-10,-8){\rotatebox{90}{\footnotesize{\cite{gu2021dro}self}}}
	\end{overpic}
	\includegraphics[width=.238\linewidth]{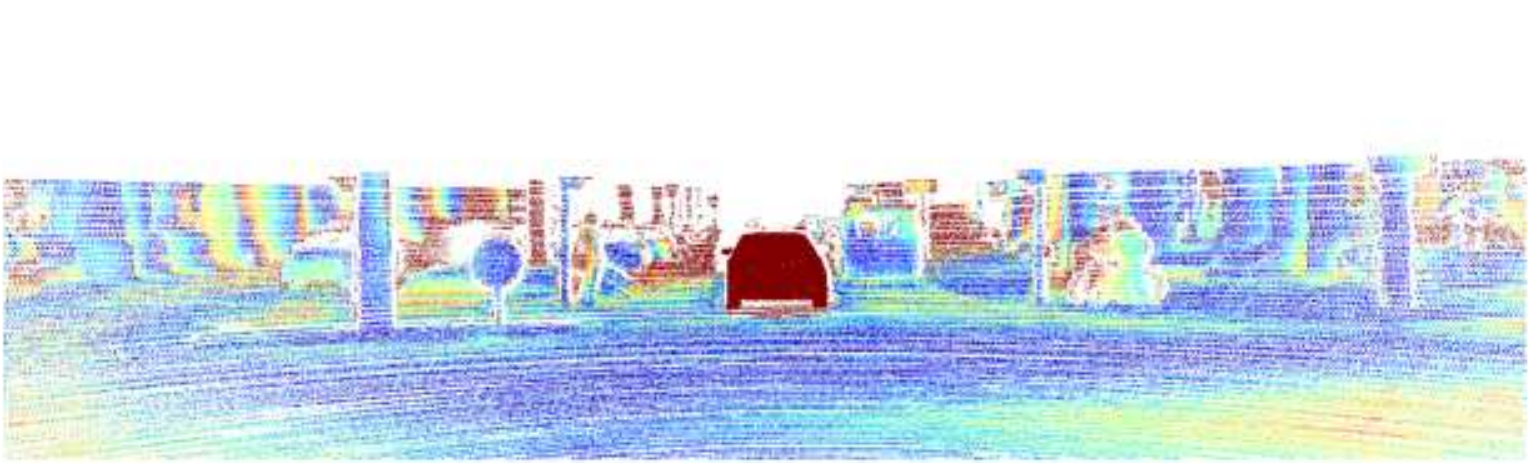}
	\includegraphics[width=.238\linewidth]{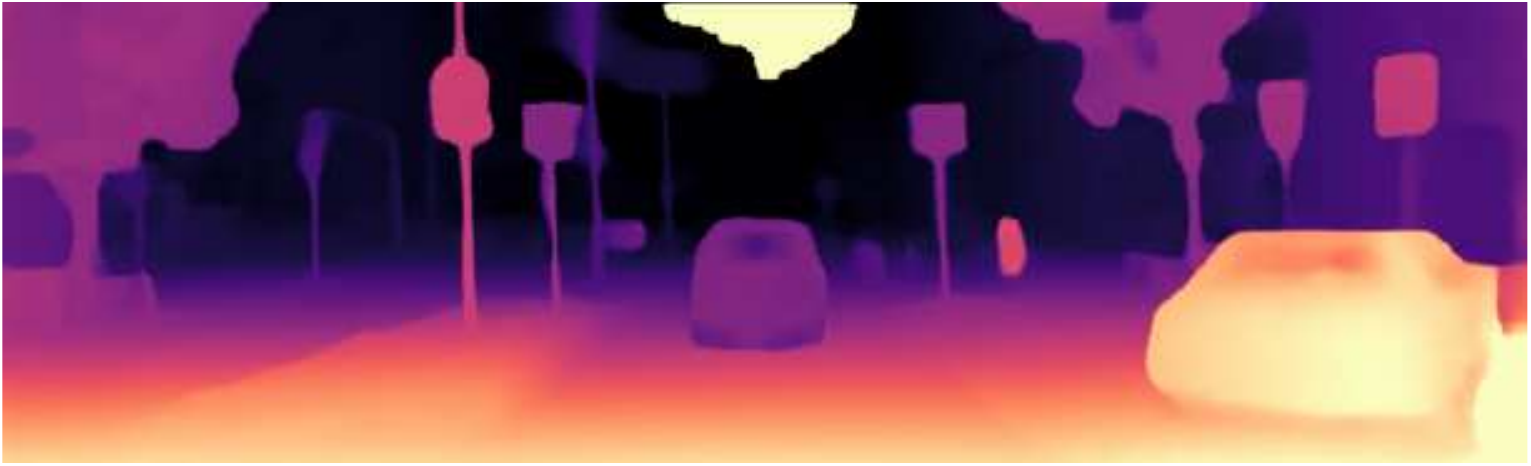}
	\includegraphics[width=.238\linewidth]{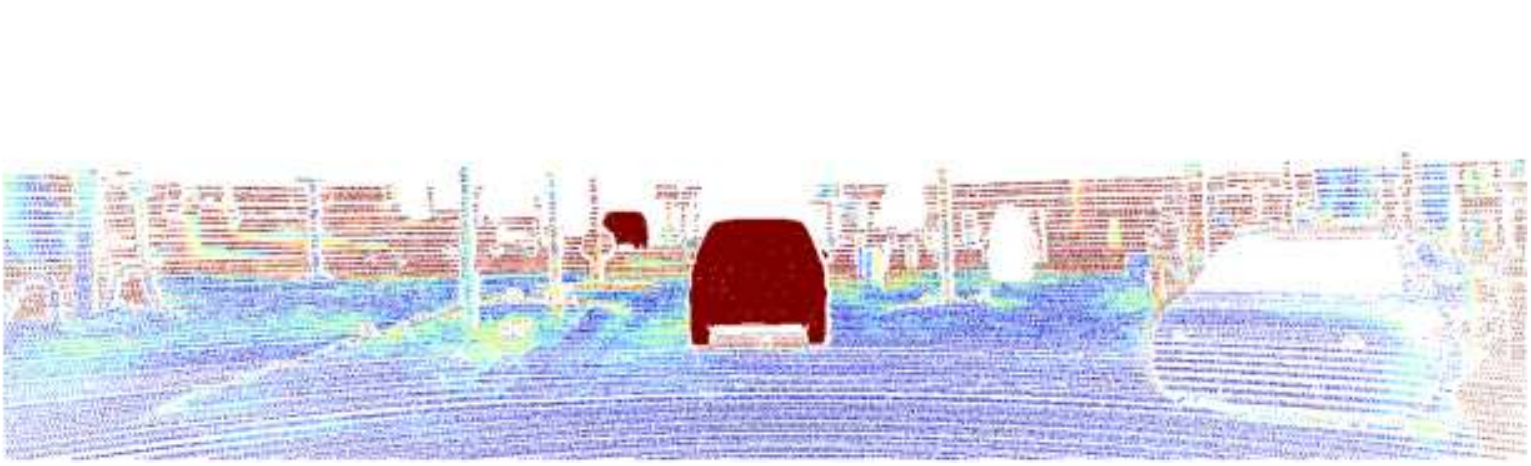}
	\\[\smallskipamount]
	
	\begin{overpic}[width=.238\linewidth, tics=10]
		{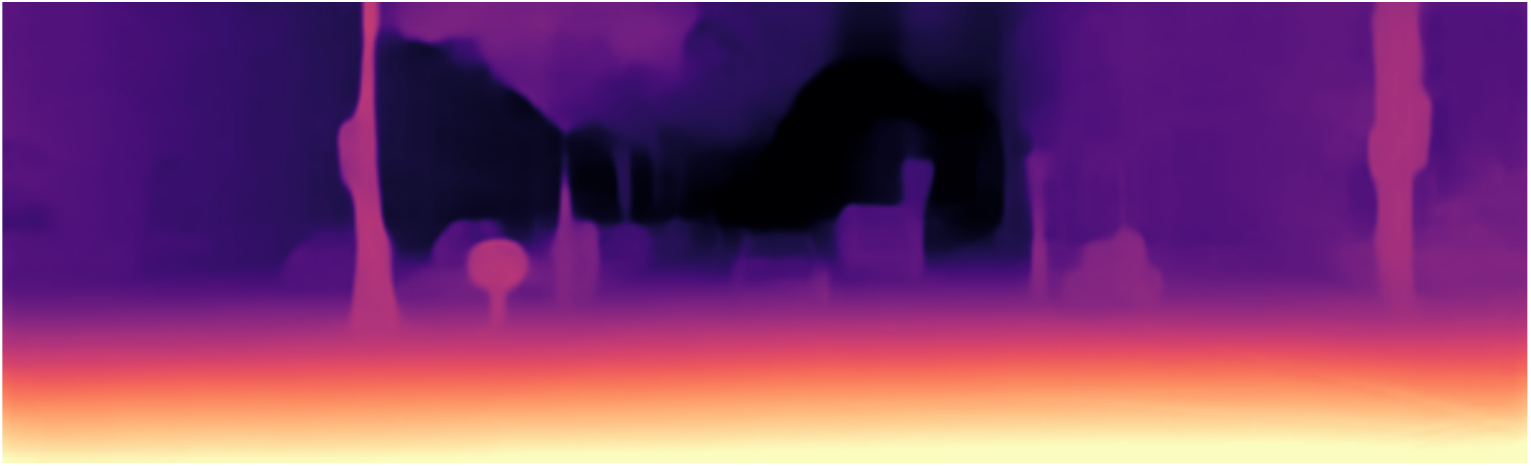}
		\put(-10,5){\rotatebox{90}{\footnotesize{\cite{watson2021temporal}}}}
	\end{overpic}
	\includegraphics[width=.238\linewidth]{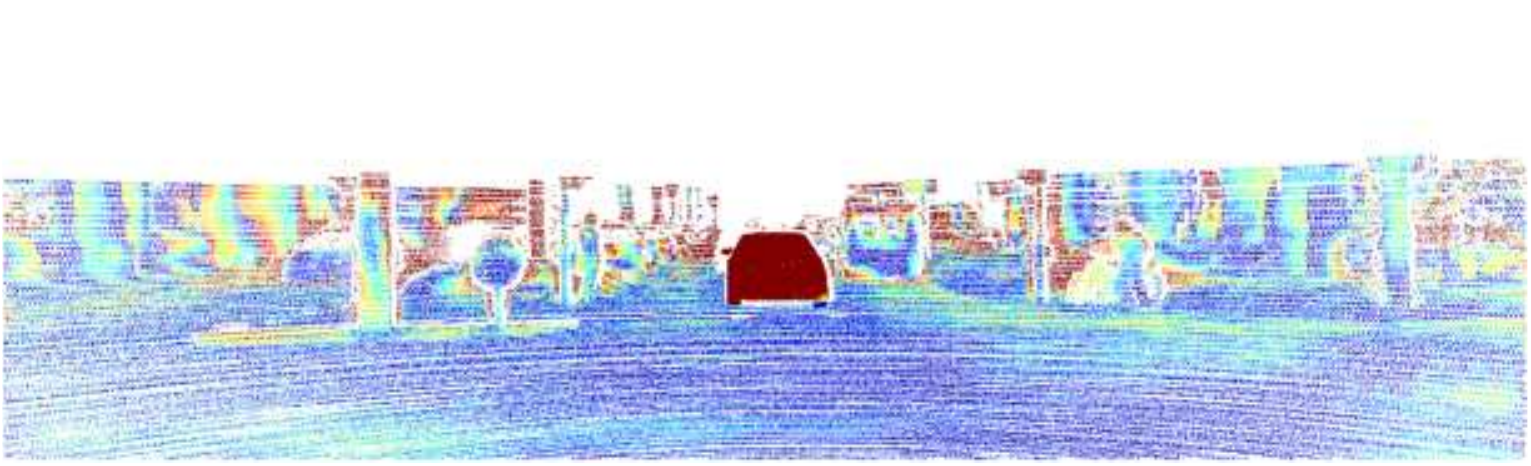}
	\includegraphics[width=.238\linewidth]{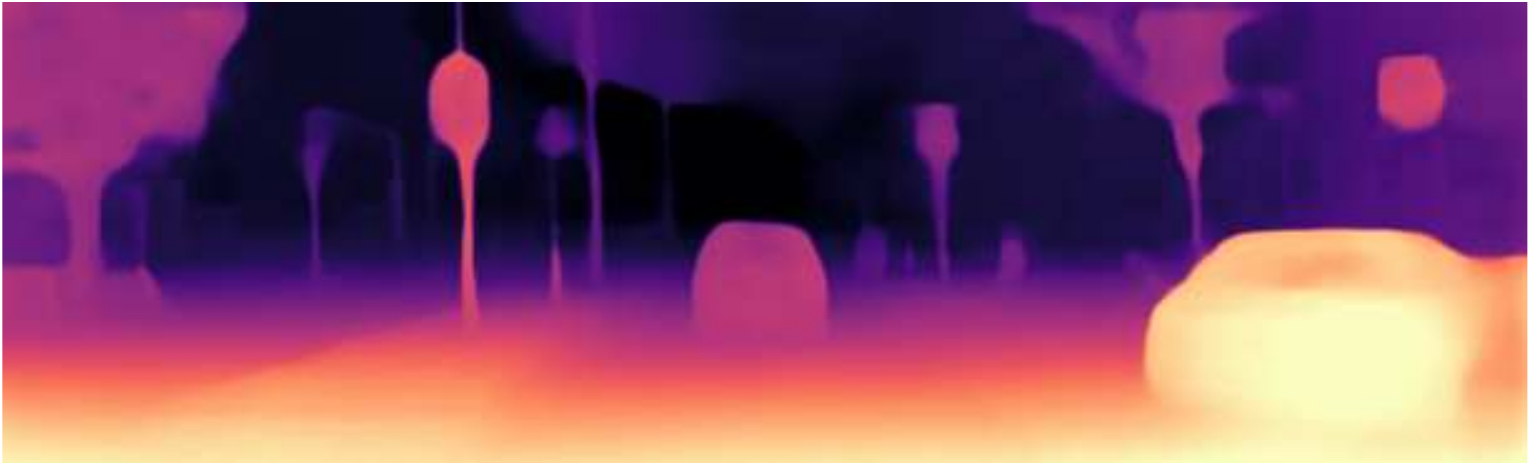}
	\includegraphics[width=.238\linewidth]{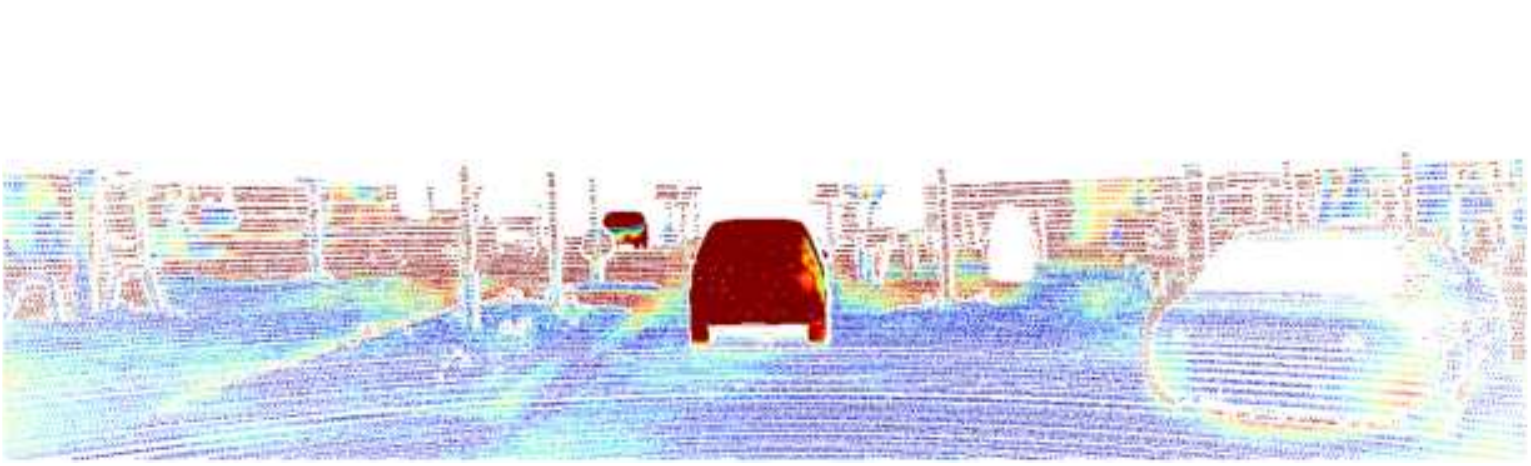}
	\\[\smallskipamount]
	\caption{Qualitative results on dense KITTI 2015 training split for 5-frame predictions. Self-supervised training struggles with objects moving colinear to the camera. While supervised frameworks do not suffer from this, they have error patterns due to sparse lidar groundtruth (e.g. cars and windows). Moving objects are usually not fully covered.}
	\label{fig:quality_depth}
\end{figure*}

\section{Training}
\label{sec:training}
We tightly follow the training protocol of \cite{teed2018deepv2d}. Each network outputs a sequence of predictions, resulting in $ \textbf{Z}_{i} $ and $ \{ \textbf{g} \}_{i} $.  
Since the model output is a sequence of $ m $ predictions the final loss is defined as the weighted sum:
\begin{align}
L &= \sum_{s=1}^{m} \gamma^{m-s} L_{s}
\end{align} with $ \gamma = 0.5 $. 
\subsection{Losses}
We experiment with a supervised and semi-supervised setting and compare to related work: 
\begin{align}
L_{spvd} &= \lambda_{1} L_{1} + \lambda_{2} L_{smooth} + \lambda_{3} L_{focal}  + \lambda_{4} L_{flow} \\
L_{semi} &= \hat{\lambda}_{1} L_{d,\; photo} + \hat{\lambda}_{2} L_{smooth} + \hat{\lambda}_{3} L_{m,\; photo} + \hat{\lambda}_{4} L_{se3} \nonumber
\end{align}

\paragraph{Supervised.}
As is common, depth is supervised with an $ l_{1} $ loss:
\begin{align}
L_{1}\left( Z,\; Z^{*} \right) &= \frac{1}{HW} \sum_{k}^{} |Z\left( \textbf{x}_{k} \right) - Z^{*}\left( \textbf{x}_{k} \right) |  
\end{align}
The camera pose graph is supervised by a reprojection error, based on the induced optical flow from the camera motion: 
\begin{align}
L_{flow}\left( \textbf{G}, \textbf{G}^{*} \right) = \frac{1}{HW} \sum_{k}^{} || & \Psi\left( G, \textbf{x}_{k}, Z\left( \textbf{x}_{k} \right) \right) - \Psi\left( \textbf{G}^{*}, \textbf{x}_{k}, Z\left( \textbf{x}_{k} \right) \right) ||_{1} \;. \nonumber
\end{align} 
Since the ground truth is sparse on real data, an additional smoothness loss is used when no ground truth is defined. The binary mask $ M $ denotes missing ground truth:
\begin{align}
	L_{smooth} = \frac{1}{HW} \sum_{k}^{} M\left( \textbf{x}_{k} \right) \odot | \partial_{x} Z\left( \textbf{x}_{k} \right) + \partial_{y} Z\left( \textbf{x}_{k} \right) |  \;.
\end{align}
It is common practice in monocular depth prediction to use edge-aware smoothing, mainly popular in unsupervised learning \cite{godard2017unsupervised,gu2021dro}:
\begin{align}
L_{smooth} = \frac{1}{HW} \sum_{k}^{} M \odot |\partial_{x} Z |  e^{-|\partial_{x} \textbf{I}|} + |\partial_{y} Z | e^{-|\partial_{y} \textbf{I}|} \;.
\end{align}
This assumes, that edges in the depth map are usually a subset of the image edges, thus sharing a spatial location. Since the loss can be readily used for supervised learning as well, we use it for both our semi-supervised and DeepV2cD experiments.

In order to train our uncertainty, we use a depth focal loss \cite{zhang2020adaptive}:
\begin{align}
L_{focal} = \frac{1}{HW} \sum_{k}^{} \sum_{d=1}^{D} \left( 1 - P^{*}\left(\textbf{x}_{k} \right) \right)^{-\delta} \cdot \left( -P\left(\textbf{x}_{k} \right) \cdot \log P^{*}\left( \textbf{x}_{k}\right) \right) ,\; \nonumber
\end{align}
with hyperparameter $ \delta $; we set $ \delta = 2 $. 
Depth supervision on real data is costly to obtain, but camera poses are often readily available. We consider a semi-supervised setting, because it can help to resolve the scale ambiguity that arises with photometric information only. We use a weighted geodesic distance on the camera poses similar to \cite{kendall2015posenet,mahendran20173d} for supervising camera pose estimation and average this over the pose graph:
\begin{align}
	L_{se3}\left(\textbf{G}^{*}, \textbf{G} \right) = d\left(\textbf{T}^{*},\textbf{T}\right) + \beta \cdot d\left( \textbf{R}^{*}, \textbf{R} \right) 
\end{align}
with $ \beta = 1.0 $.
\paragraph{Self-supervised.}
Self-supervised learning uses the image data for creating a training signal. Based on the estimated depth $ Z $ and a relative pose $ \textbf{g} $, a warped image $ \textbf{I}^{\prime} $ can be constructed with $ \Psi $ from the input image $ \textbf{I} $ for each key-reference image pair. The similarity is measured by a weighted sum of structural similarity index (SSIM) and $ L1 $ loss \cite{godard2017unsupervised,godard2019digging,guizilini20203d}:
\begin{align}
	L_{photo}\left( \textbf{I}_{i},\; \textbf{I}_{i}^{\prime} \right) &= \alpha  \frac{1 - SSIM\left( \textbf{I}_{i},\; \textbf{I}_{i}^{\prime}\right)}{2} + \left( 1 - \alpha \right) || \textbf{I}_{i} - \textbf{I}_{i}^{\prime} ||_{1} \; .
\end{align}
We adopt the minimum fusion strategy from \cite{godard2019digging} for fusing the photometric losses across view pairs. Similar to the supervised case, we can enforce edge-aware smoothness for the depth predictions, but with $ M $ being all in-view pixels.
\paragraph{Weighting.}
We adopt hyperparameters for balancing the original loss terms from \cite{teed2018deepv2d} and determine others empirically. For our results, we set $ \lambda = \left[ 1.0, 0.02, 0.002,1.0 \right] $ and $ \hat{\lambda} = \left[10.0, 0.02, 10.0, 1.0 \right] $.

\subsection{Implementation details}
We extend on the published code of Teed et al. \cite{teed2018deepv2d}, but upgrade their implementation to \textit{Tensorflow 2} \cite{abadi2016tensorflow}. All components from the network are trained from scratch with RMSProp \cite{tieleman2012lecture} if not stated otherwise. We use standard color and flip augmentation similar to \cite{teed2018deepv2d,gu2021dro} and small random camera pose perturbation as \cite{teed2018deepv2d}. Our input and output resolution is 192 $ \times $ 1088, which is reasonable for the limited lidar coverage. We use $ n=5 $ frames as input as this turns out to be enough temporal information to achieve optimal performance, see \cite{teed2018deepv2d}.   

Training can be divided into two stages: 1\@.$ \, $ the motion module is trained alone with interpolated groundtruth depth as input. 2\@.$ \, $the motion network gets a cached depth prediction with 
increasing likelihood from the previous epoch as input instead of the depth groundtruth and the depth network takes the predicted camera poses. Both networks are trained jointly on the combined loss and 
therefore depend on each other already during training. We train for approx\@.$\,\left[ 5, 15 \right] $ epochs with batch sizes $ \left[ 12,3 \right] $ for the respective stages like the baseline from \cite{teed2018deepv2d} if not stated otherwise. During inference we run 5 iterations of the networks compared to 8 in the reference implementation, because we found that this already matches the reported performance.

\section{Experiments}
\label{seq:experiments}
We test DeepV2cD and our own \textit{Tensorflow 2} DeepV2D implementation on multiple automotive datasets that include dynamic objects. Our primary focus lies on the depth prediction performance and we include both multi-view and single-view comparisons. 
As common in deep monocular multi-view SfM we report results for the ground truth scale aligned depth \cite{ummenhofer2017demon,tang2018ba,teed2018deepv2d}. We compare with current SotA multi-view frameworks and report results for their publicly available models.

\subsection{Datasets}
\paragraph{KITTI.}
The KITTI dataset \cite{geiger2012we} contains image and lidar sequences from a moving car and is widely established for evaluating depth estimation. We follow the Eigen train/test split and evaluation protocol \cite{eigen2014depth} with the official ground truth maps. The original dataset has only sparse depth ground truth and misses moving object labels. To adress these shortcomings, the KITTI 2015 dataset was published \cite{menze2015object} with dense ground truth and labels for 200 frames. We evaluate on the KITTI 2015 training split to show the performance gap between the static scene and dynamic objects. During evaluation we also use the Eigen crop as on KITTI 2012.

\paragraph{Virtual KITTI 2.}
Virtual KITTI \cite{cabon2020virtual} is a synthetic dataset, consisting of several driving sequences from the real KITTI dataset imitated with the unity engine. It contains diverse weather and lighting conditions and moving objects. Even though it does not cover as many driving scenes as the original real data, we use it to investigate the zero-shot generalization performance of our backbone network DeepV2D and improved performance for our supervised DeepV2cD model. We generated our own 95/5 train/test split in similar fashion to the Eigen split. We noticed, that some frames have misaligned ground planes and objects that appear/vanish in the middle of the scene. By removing such frames and ones with no camera motion, we achieve a split of approx. 14.3k and 754 train/test frames. We train for $ \left[9, 20\right] $ epochs and we train sequentially when combining multiple datasets. We believe, that with a better domain adaption strategy \cite{guizilini2021geometric} results can be improved further than sequential training on the multiple datasets. 

\paragraph{Cityscapes.}
The Cityscapes dataset \cite{cordts2016cityscapes} is an autonomous driving dataset collected from various cities in Germany. The dataset contains many moving object classes, such as cars and pedestrians. It is mainly used for object detection, but also contains 1500 frames with Semi-Global Matching (SGM) \cite{hirschmuller2011semi} depth ground truth. We evaluate on this dataset to test zero-shot generalization performance and follow the evaluation protocol of \cite{casser2019unsupervised,watson2021temporal}.

\begin{table*}[h!]
	\caption{Results on KITTI Eigen split with improved ground truth. Multi-view frameworks have higher accuracy than single-view. We achieve the best results when regularizing the cost volume of our backbone network and training with more data.}
	\label{tab:kitti12}
	\resizebox{\textwidth}{!}{%
		\begin{tabular}{lccccccccc}
			\toprule
			Method & Views & Supervised & $ W \times H $ & Abs Rel $ \downarrow $ & Sq Rel $ \downarrow $ & RMSE $ \downarrow $ & RMSE $ \log $ $ \downarrow $ & $ \delta < 1.25 $ $ \uparrow $ & $ \delta < 1.25^{2} $ $ \uparrow $ \\
			\midrule
			MonoDepth \cite{godard2019digging} & 1 & \xmark & $ 1024 \times 320 $ & 0.091 & 0.531 & 3.742 & 0.135 & 0.916 & 0.984 \\
			Kuznietsov et al. \cite{kuznietsov2017semi} & 1 & \cmark & $ 621 \times 187 $ & 0.089 & 0.478 & 3.610 & 0.138 & 0.906 & 0.98 \\
			Packnet-SfM \cite{guizilini20203d} & 1 & (\cmark) & $ 640 \times 192 $ & 0.078 & 0.420 & 3.485 & 0.121 & 0.931 & 0.986 \\
			DORN \cite{fu2018deep} & 1 & \cmark & $ 513 \times 385 $  & 0.069 & 0.300 & 2.857 & 0.112 & 0.945 & 0.988 \\
			\midrule
			BANet \cite{tang2018ba} & 5 & \cmark & - & 0.083 & - & 3.640 & 0.134 & - & - \\
			Manydpepth \cite{watson2021temporal} & 2 & \xmark & $ 1024 \times 320 $ & 0.055 & 0.313 & 3.035 & 0.094 & 0.958 & 0.990 \\
			Manydpepth \cite{watson2021temporal} & 5 & \xmark & $ 1024 \times 320 $ & 0.055 & 0.312 & 3.034 & 0.094 & 0.958 & 0.991 \\
			DRO \cite{gu2021dro} & 2 & \xmark & $ 960 \times 320 $ & 0.057 & 0.342 & 3.201 & 0.123 & 0.952 & 0.989 \\
			DRO \cite{gu2021dro} & 5 & \xmark & $ 960 \times 320 $ & 0.064 & 0.381 & 3.262 & 0.120 & 0.951 & 0.988 \\
			DRO \cite{gu2021dro} & 2 & \cmark & $ 960 \times 320 $ & 0.046 & 0.210 & 2.674 & 0.083 & 0.969 & \underline{0.993} \\
			DRO \cite{gu2021dro} & 5 & \cmark & $ 960 \times 320 $ & 0.047 & 0.212 & 2.711 & 0.084 & 0.968 & \textbf{0.994} \\
			DeepV2D (ours) (VK) & 5 & \xmark & $ 1088 \times 192 $ & 0.060 & 0.423 & 3.302 & 0.110 & 0.950 & 0.984 \\
			DeepV2D (ours) (VK + K) & 5 & (\cmark) & $ 1088 \times 192 $ & 0.058 & 0.669 & 3.246 & 0.100 & 0.960 & 0.985 \\
			DeepV2D \cite{teed2018deepv2d} & 2 & \cmark &  $ 1088 \times 192 $ & 0.064 & 0.350 & 2.946 & 0.120 & 0.946 & 0.982 \\
			DeepV2D \cite{teed2018deepv2d} & 5 & \cmark &  $ 1088 \times 192 $ & \underline{0.037} & 0.174 & 2.005 & 0.074 & 0.977 & \underline{0.993} \\
			DeepV2cD (ours) & 5 & \cmark & $ 1088 \times 192 $ & \underline{0.037} & \underline{0.167} & \underline{1.984} & \underline{0.073} & \underline{0.978} & \textbf{0.994} \\
			DeepV2cD* (ours) (K + VK) & 5 & \cmark & $ 1088 \times 192 $ & \textbf{0.035} &\textbf{ 0.158} & \textbf{1.877} & \textbf{0.071} & \textbf{0.980} & \textbf{0.994} \\
			\bottomrule
		\end{tabular}
	}
\end{table*}

\subsection{Accuracy of multi-view networks}
Table \ref{tab:kitti12} shows the current SotA for monocular depth prediction on the KITTI Eigen split. It can be seen that DeepV2D and DeepV2cD achieve the best reconstruction accuracy in the 5-view setting on the KITTI Eigen split. All multi-view frameworks achieve better results than single-view ones. While DRO \cite{gu2021dro} and Manydepth \cite{watson2021temporal} are significantly faster than DeepV2D and DeepV2cD, they are less accurate in a 5-view setting. Depending on the training, not all multi-view models achieve higher accuracy with more views added compared to the 2-view setting. Our cost volume regularization and uncertainty strategy gives better results than the DeepV2D baseline \cite{teed2018deepv2d} while needing fewer inference iterations.

All supervised multi-view frameworks except DeepV2cD can be trained with self-supervision as well. We did not manage to train DeepV2D in this setting. However, we can show, that it achieves good performance by simply training on the small Virtual KITTI (VK) dataset. Current unsupervised multi-view approaches perform best when trained on the target dataset. The gap can be further closed after finetuning semi-supervised (\cmark) without expensive depth ground truth for just 5 epochs. Manydepth utilizes a test-time refinement, which we did not experiment with for DRO and DeepV2D. 

\begin{table}[h!]
	\caption{Results on KITTI 2015 train split. Metrics are slightly worse for all frameworks compared to the KITTI 2012 Eigen split.}
	\label{tab:kitti15}
	\centering
		\begin{tabular}{lccccc}
			\toprule
			Method & Supervised & ARE Dyn. $ \downarrow $ & ARE Static $ \downarrow $ & ARE All $ \downarrow $ & $ \delta < 1.25 $ $ \uparrow $ \\
			\midrule
			DRO \cite{gu2021dro} & \xmark & 449.1 & 0.081 & 0.276 & 0.795 \\
			Manydpepth & \xmark & 0.177 & 0.0641 & 0.090 & 0.914 \\
			DRO \cite{gu2021dro} & \cmark & \textbf{0.069} & 0.050 & \textbf{0.056} & 0.953 \\
			DeepV2D (ours) & \cmark & 0.152 & \underline{0.042} & 0.071 & \underline{0.959} \\
			DeepV2cD* (ours) & \cmark & \underline{0.127} & \textbf{0.039} & \underline{0.062} & \textbf{0.965} \\
			\bottomrule
	\end{tabular}
\end{table}

\begin{table*}
	\caption{Zero shot cross-dataset generalization of multi-view frameworks. We evaluate models trained on KITTI (K) on Cityscapes (CS).}
	\label{tab:cityscapes}
	\resizebox{\textwidth}{!}{%
		\begin{tabular}{lccccccccc}
			\toprule
			Method & Supervised & Dataset & Abs Rel $ \downarrow $ & Sq Rel $ \downarrow $ & RMSE $ \downarrow $ & RMSE $ \log $ $ \downarrow $ & $ \delta < 1.25 $ $ \uparrow $ & $ \delta < 1.25^{2} $ $ \uparrow $ & $ \delta < 1.25^{3} $ $ \uparrow $ \\
			\midrule
			Manydpepth \cite{watson2021temporal} & \xmark & CS & 0.114 & 1.193 & 6.223 & 0.170 & 0.875 & 0.967 & 0.989 \\
			\midrule
			DRO \cite{gu2021dro} & \xmark & K $ \rightarrow $ CS & 0.328 & 7.348 & 11.656 & 0.597 & 0.548 & 0.747 & 0.822 \\
			DRO \cite{gu2021dro} & \cmark & K $ \rightarrow $ CS & 0.157 & 2.228 & 10.306 & 0.299 & 0.777 & 0.900 & 0.948 \\
			Manydpepth \cite{watson2021temporal} & \xmark & K $ \rightarrow $ CS & 0.162 & 1.697 & 8.230 & 0.229 & 0.764 & 0.935 & \textbf{0.979} \\
			DeepV2D & \cmark & K $ \rightarrow $ CS & \underline{0.109} & \underline{1.479} & \underline{6.7633} & \underline{0.1842} & \underline{0.876} & \underline{0.952} & \underline{0.978} \\
			DeepV2cD* & \cmark & K + VK $ \rightarrow $ CS & \textbf{0.104} & \textbf{1.325} & \textbf{6.7328} & \textbf{0.1792} &\textbf{ 0.883} & \textbf{0.955} & \textbf{0.979} \\
			\midrule
			DeepV2cD* filtered $ 80 $\% & \cmark & K + VK $ \rightarrow $ CS & 0.070 & 0.559 & 4.188 & 0.124 & 0.930 & 0.975 & 0.989 \\
			\bottomrule
	\end{tabular}
	}
\end{table*}

\begin{figure*}[h!]
	\centering
	\captionsetup{skip=10pt}
	\includegraphics[width=.25\linewidth]{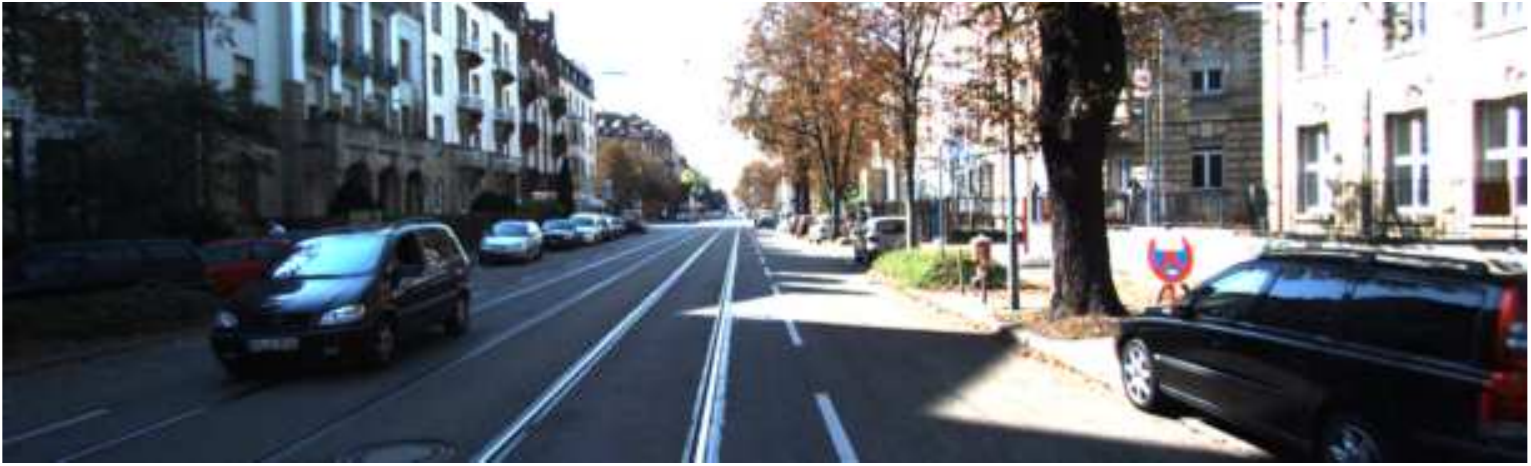}\hfill
	\begin{overpic}[width=.25\linewidth, tics=10]
		{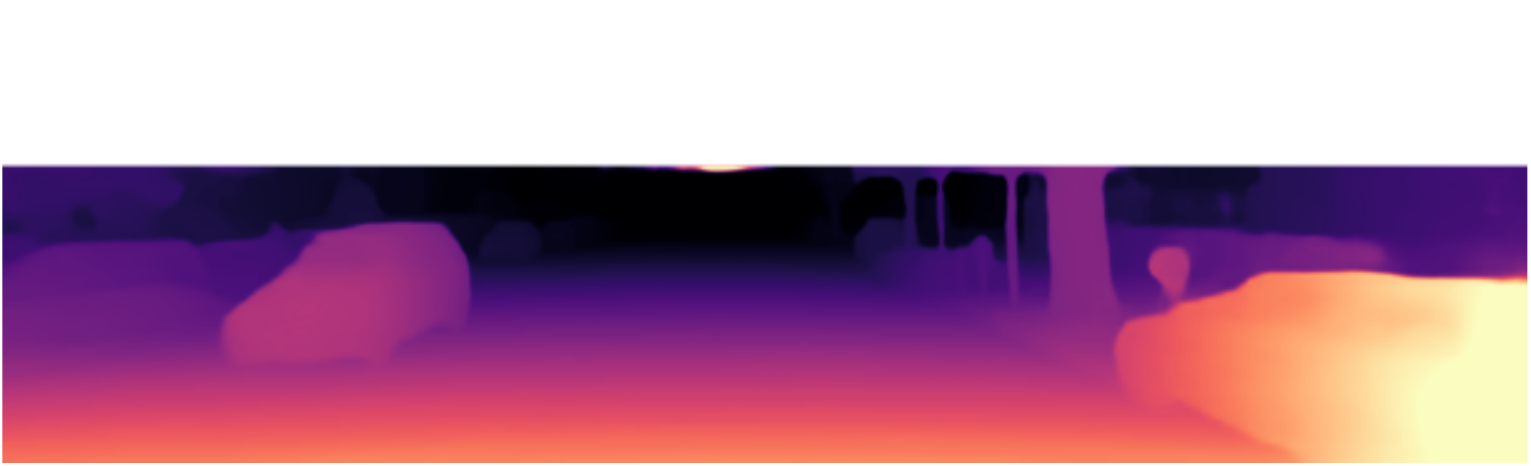}
		\put(17, 25){Depth prediction}
	\end{overpic}\hfill
	\begin{overpic}[width=.25\linewidth, tics=10]
		{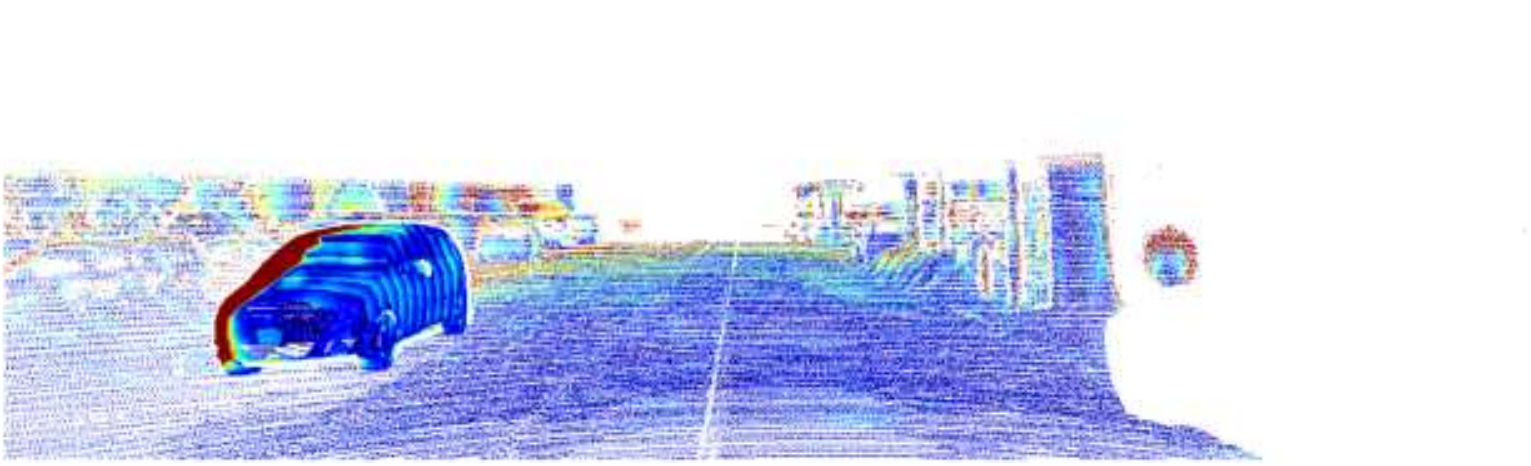}
		\put(22, 25){Abs. rel. error}
	\end{overpic}\hfill
	\begin{overpic}[width=.25\linewidth, tics=10]
		{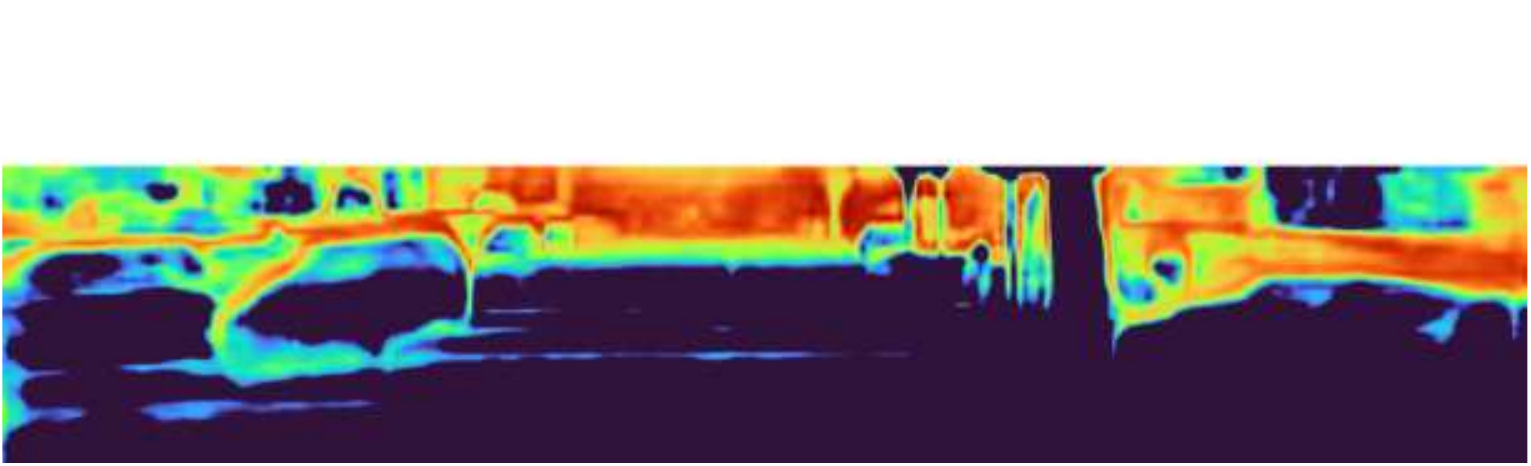}
		\put(27, 25){Uncertainty}
	\end{overpic}\hfill
	\\[\smallskipamount]
	\includegraphics[width=.25\linewidth]{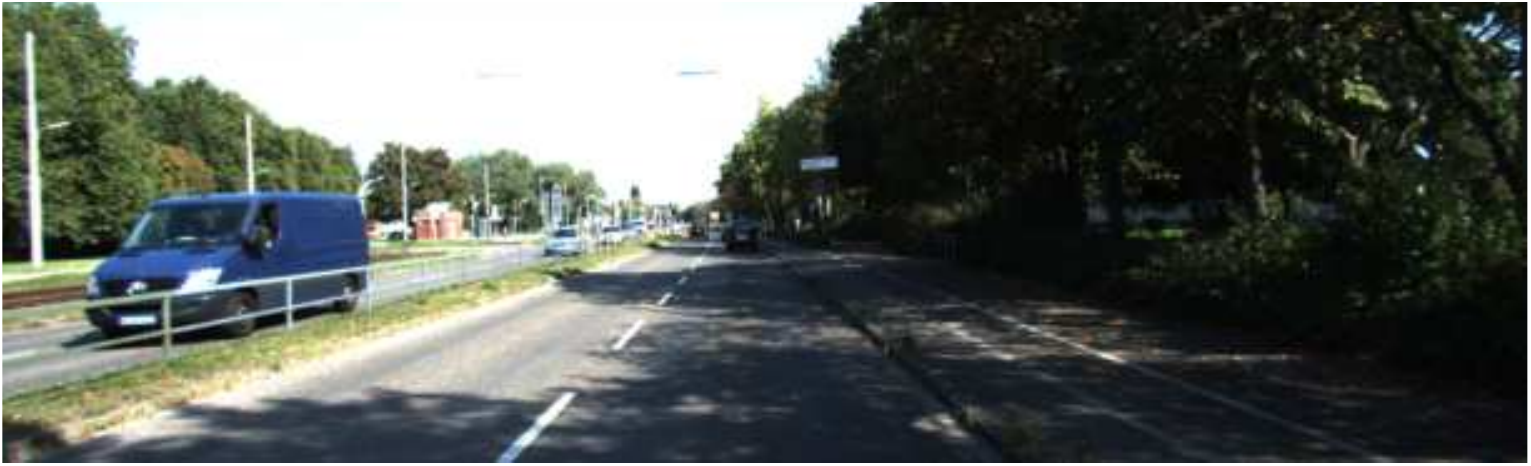}\hfill
	\includegraphics[width=.25\linewidth]{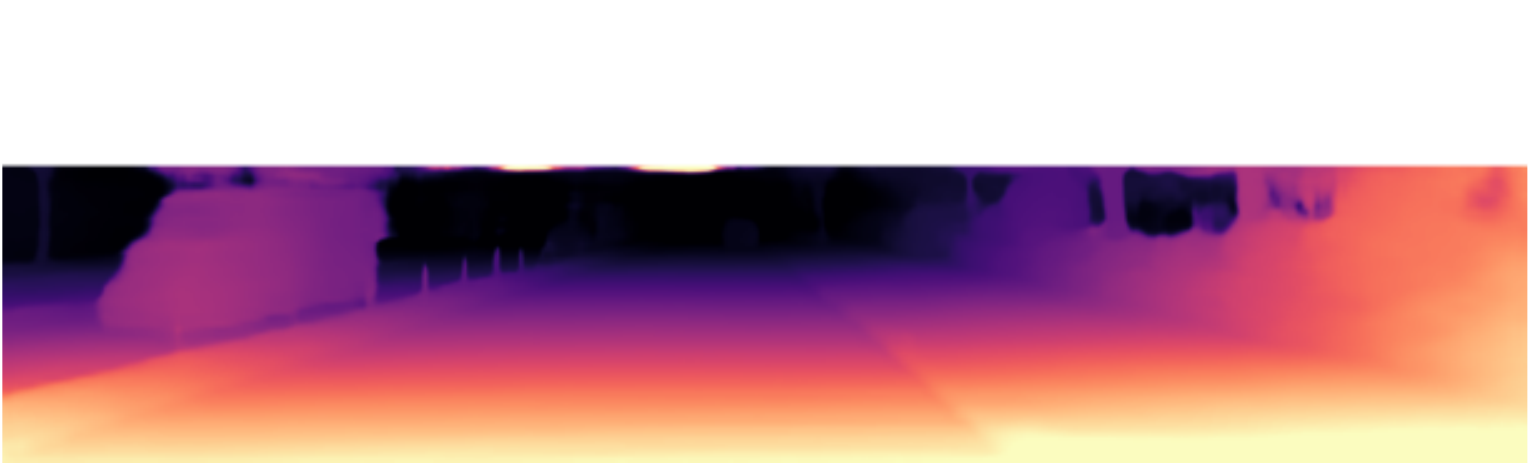}\hfill
	\includegraphics[width=.25\linewidth]{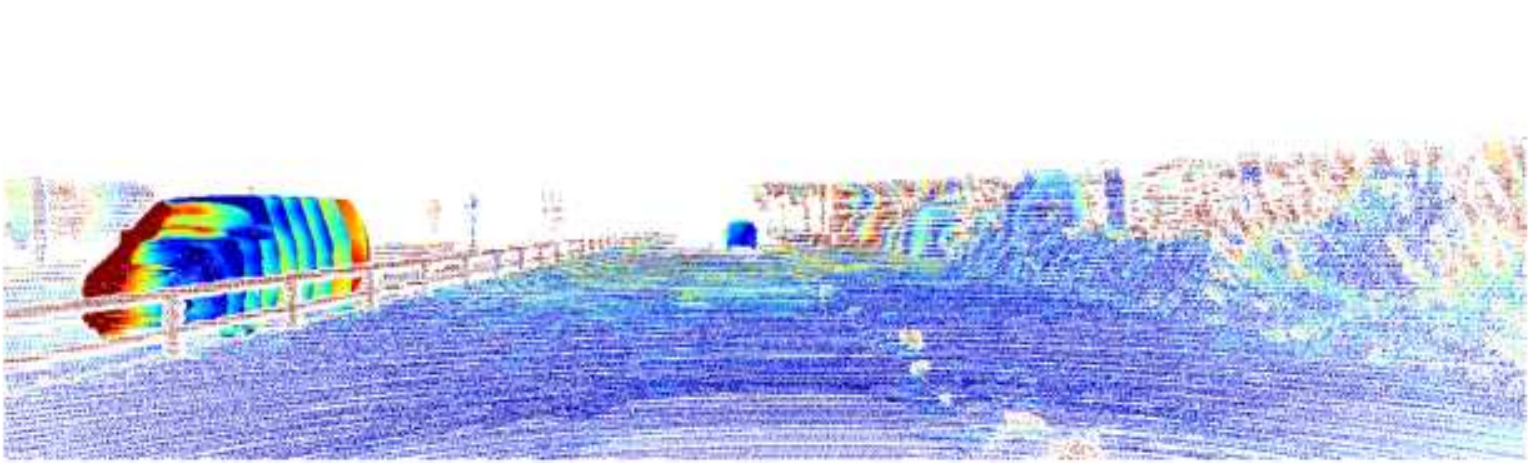}\hfill
	\includegraphics[width=.25\linewidth]{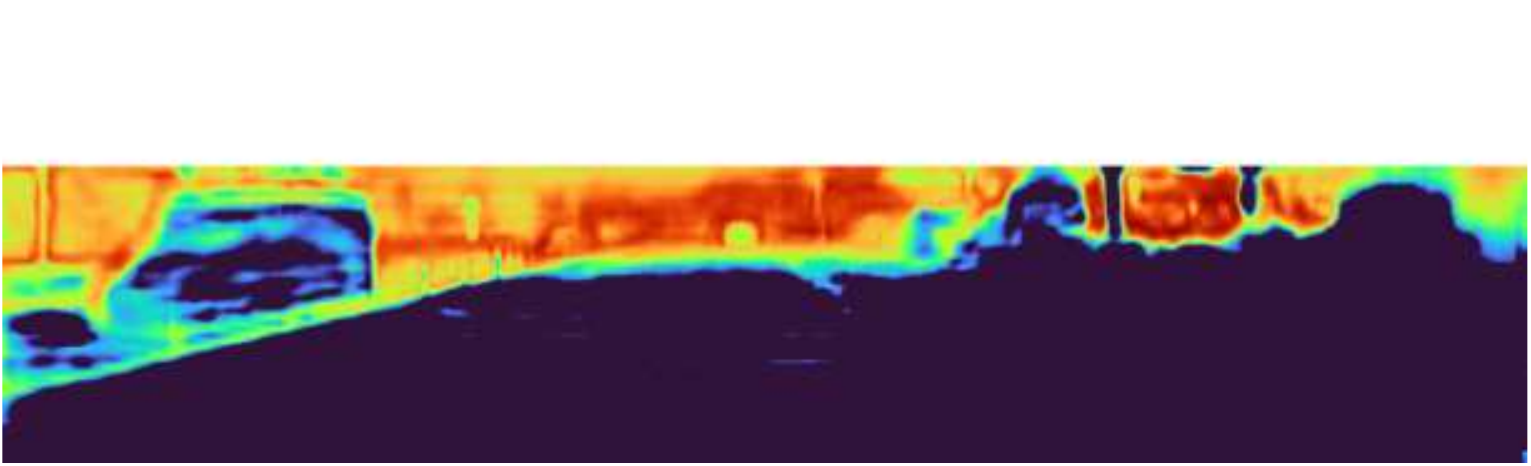}
	\\[\smallskipamount]
	\includegraphics[width=.25\linewidth]{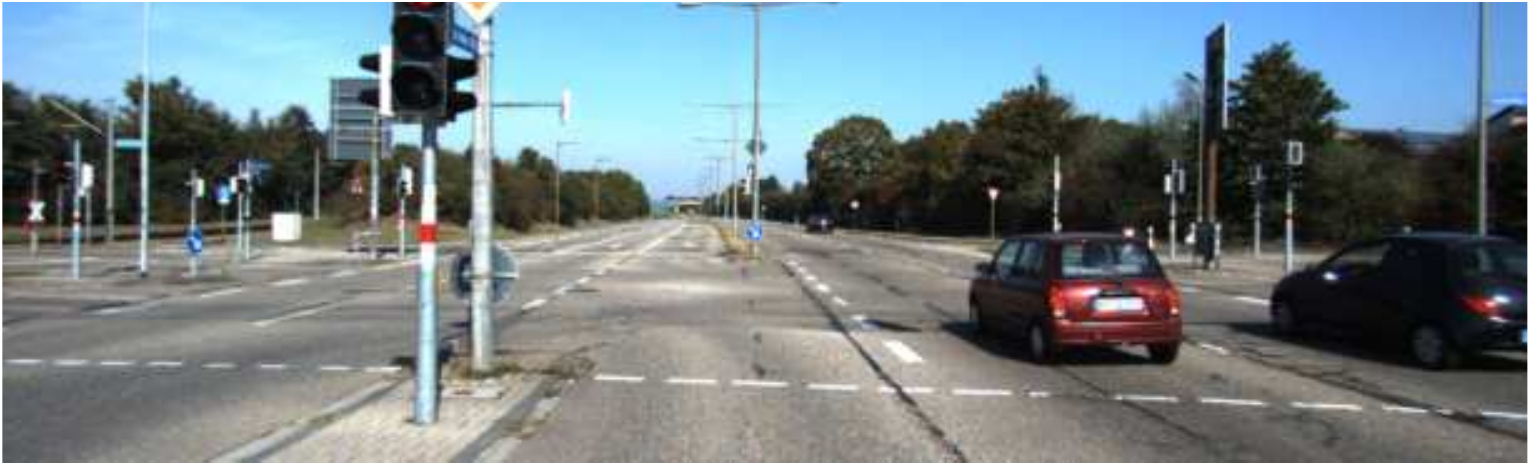}\hfill
	\includegraphics[width=.25\linewidth]{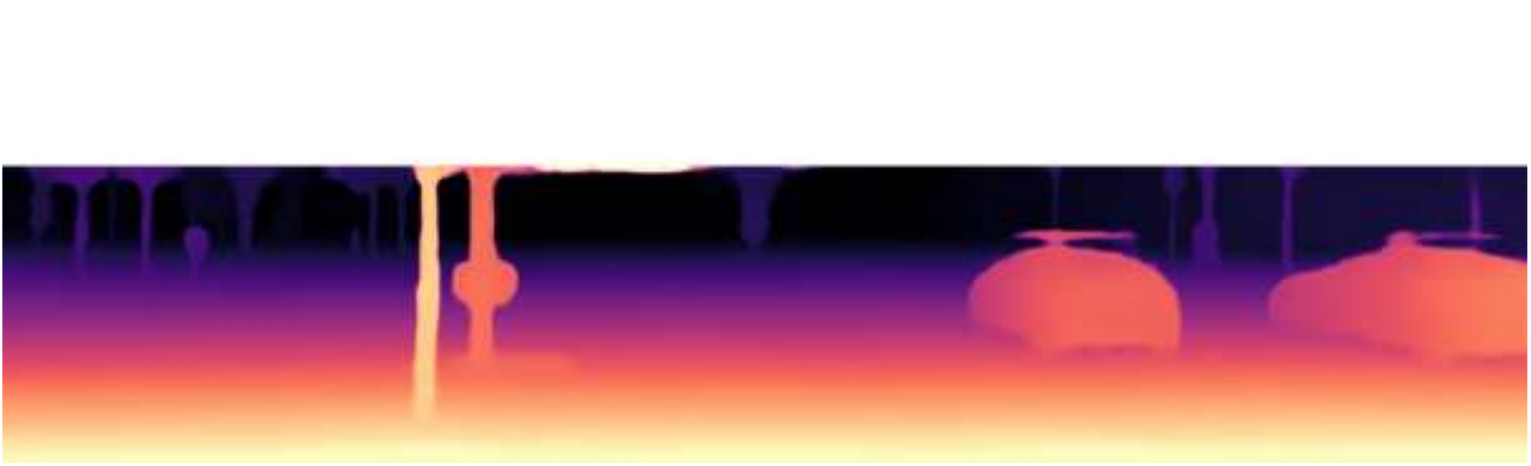}\hfill
	\includegraphics[width=.25\linewidth]{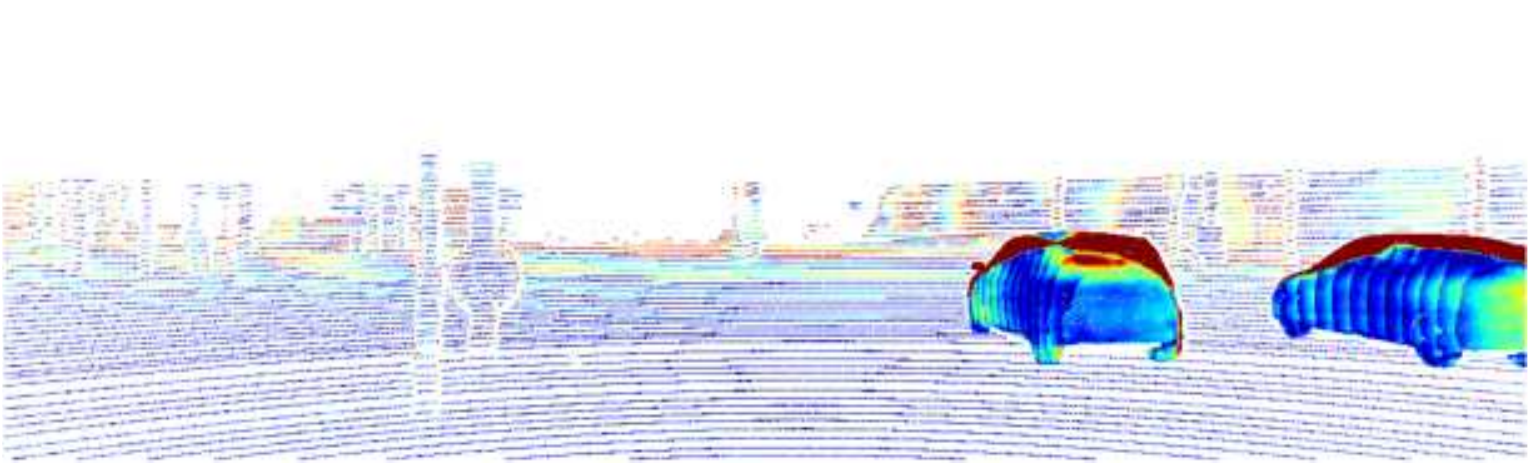}\hfill
	\includegraphics[width=.25\linewidth]{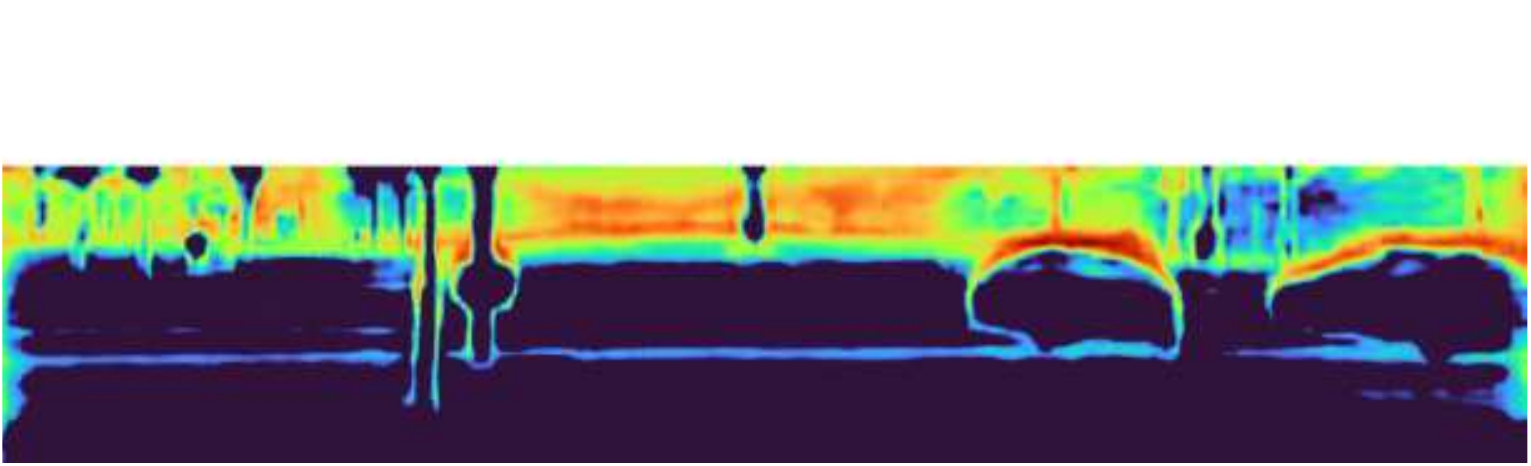}
	\caption{Qualitative results on three samples of KITTI 2015 training split. Warm colors indicate high error and uncertainty. DeepV2cD makes errors in areas of occlusion, missing ground truth, reflective surfaces or thin structures. The uncertainty head learns to predict these areas reliably with no overhead. Most problematic for reconstruction of dynamic objects is missing lidar supervision, noticable by the cut off roofs and holes at windows. DeepV2cD reconstructs these regions as part of the background. The uncertainty is strongest in these areas and in the back of the scene at far depths.}
	\label{fig:uncertainty}
	\vspace{-10pt}
\end{figure*}

Table \ref{tab:kitti15} shows results on the KITTI 2015 training split to emphasize the difference between static and dynamic scene. In Figure \ref{fig:quality_depth}, we show some qualitative examples. DeepV2cD achieves the best reconstruction of the static scene. All frameworks perform systematically worse on moving objects. Unsupervised training cannot resolve the inherent ambiguity resulting from the independent object motion and thus results in gross errors when not taking care of regularization. 
Manydepth resolves this problem partially by regularizing their network with a single-view teacher network compared to the self-supervised DRO during training. 
We observe, that even though no supervised framework addresses the object motion directly in their reprojection operator, the reconstruction works in most cases. We believe, that this is due to cost volume regularization in DeepV2D and DeepV2cD and the temporal information of the recurrent network in DRO. Since we could not observe these errors on virtual data, we believe that these errors are due to missing supervision with the sparse lidar. While the systematic model errors of DeepV2cD and DRO on the dynamic parts look qualitatively similar, DRO has a significantly lower gap to the static reconstruction. We observe, that DeepV2cD sometimes does not reconstruct the car shapes as well and produces more gross outliers. A possible explanation for this is the longer training time of DRO compared to DeepV2cD (50 epochs compared to 15) and the pre-trained ImageNet \cite{russakovsky2015imagenet} weights. Pretraining on Virtual KITTI does not resolve this issue completely.

Table \ref{tab:cityscapes} shows the zero-shot cross-dataset generalization performance of current SotA multi-view frameworks. DeepV2D and DeepV2cD can generalize well to other datasets. We generalize best from KITTI to Cityscapes and outperform unsupervised single-view networks trained on the target. This could be explained with the explicit geometric pose graph optimization for motion estimation and the cost volume. The learned uncertainty transfers well to Cityscapes. When keeping $ 80\% $ of the pixels, we effectively reach $ 50 \% $ the error of other frameworks in this setting. 

\begin{figure*}[h!]
	\begin{subfigure}{.49\textwidth}
		\centering
		\includegraphics[width=0.95\linewidth]{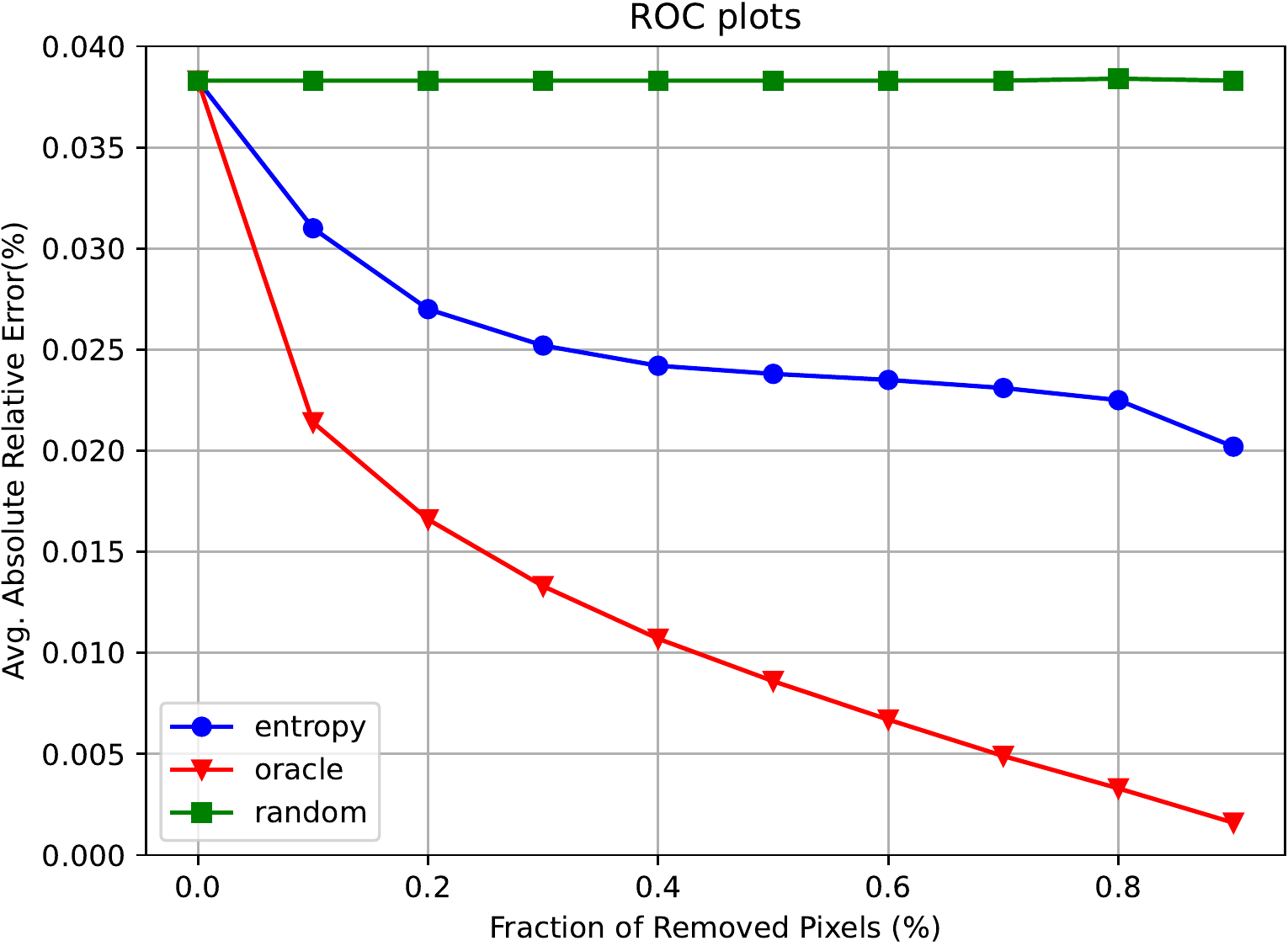}
		\caption{DeepV2D on KITTI 2012 Eigen test split}
	\end{subfigure}%
	\begin{subfigure}{.49\textwidth}
		\centering
		\includegraphics[width=0.95\linewidth]{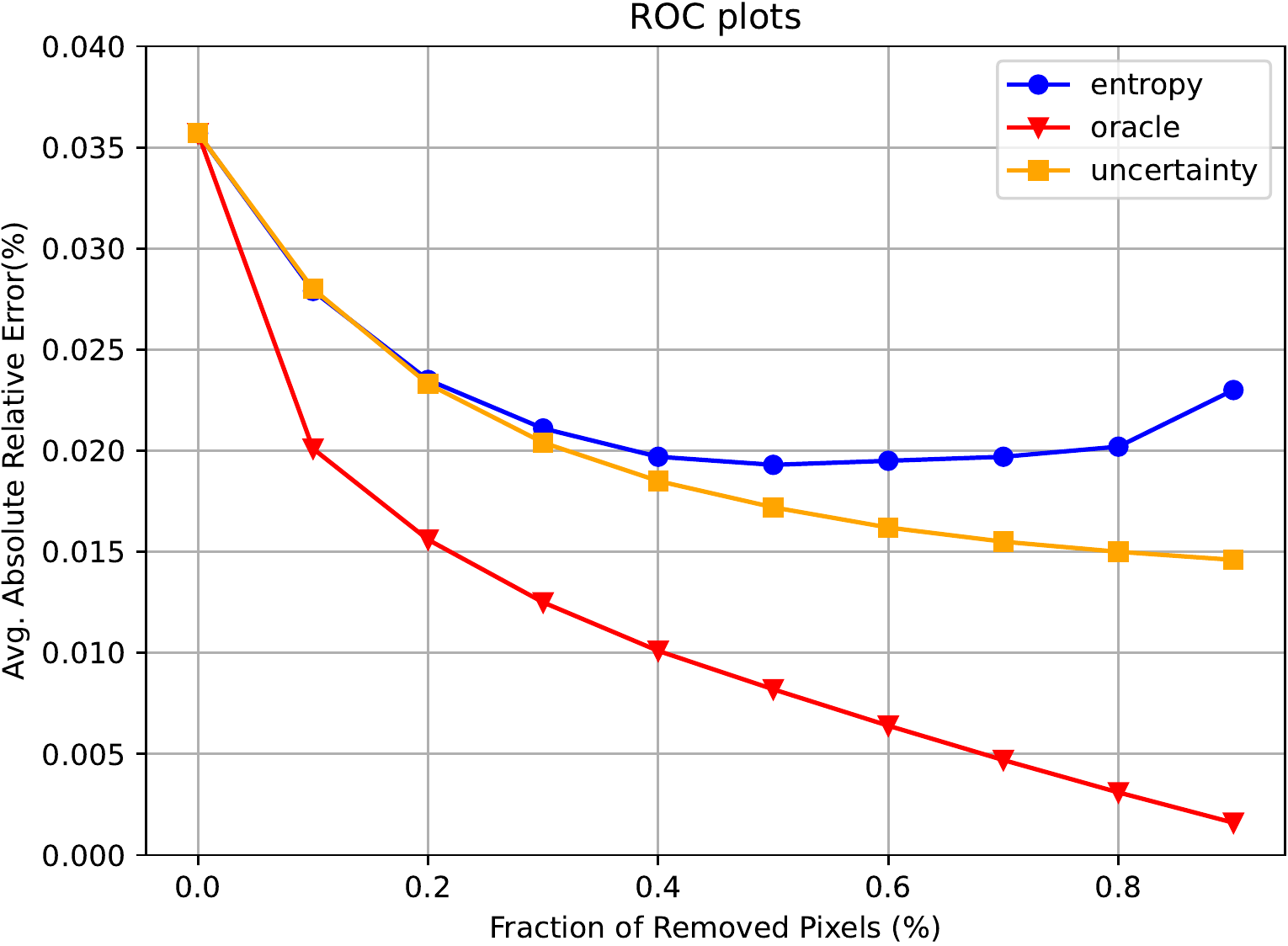}
		\caption{DeepV2cD* on KITTI 2012 Eigen test split}
	\end{subfigure}\hfill
	\label{fig:kitti12_roc}

	\begin{subfigure}{.49\textwidth}
		\centering
		\includegraphics[width=0.95\linewidth]{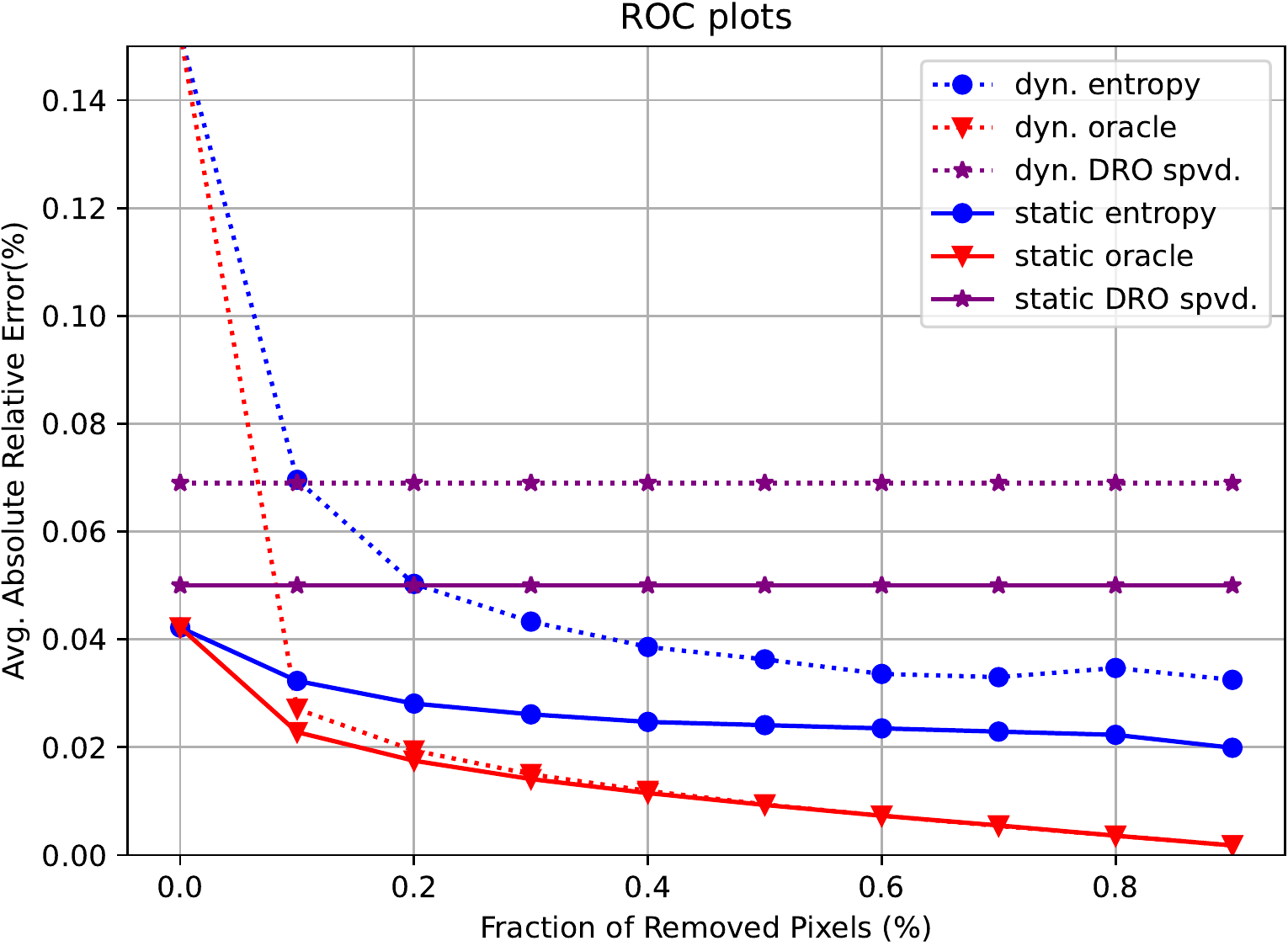}
		\caption{DeepV2D on KITTI 2015 train split}
	\end{subfigure}%
	\begin{subfigure}{.49\textwidth}
		\centering
		\includegraphics[width=0.95\linewidth]{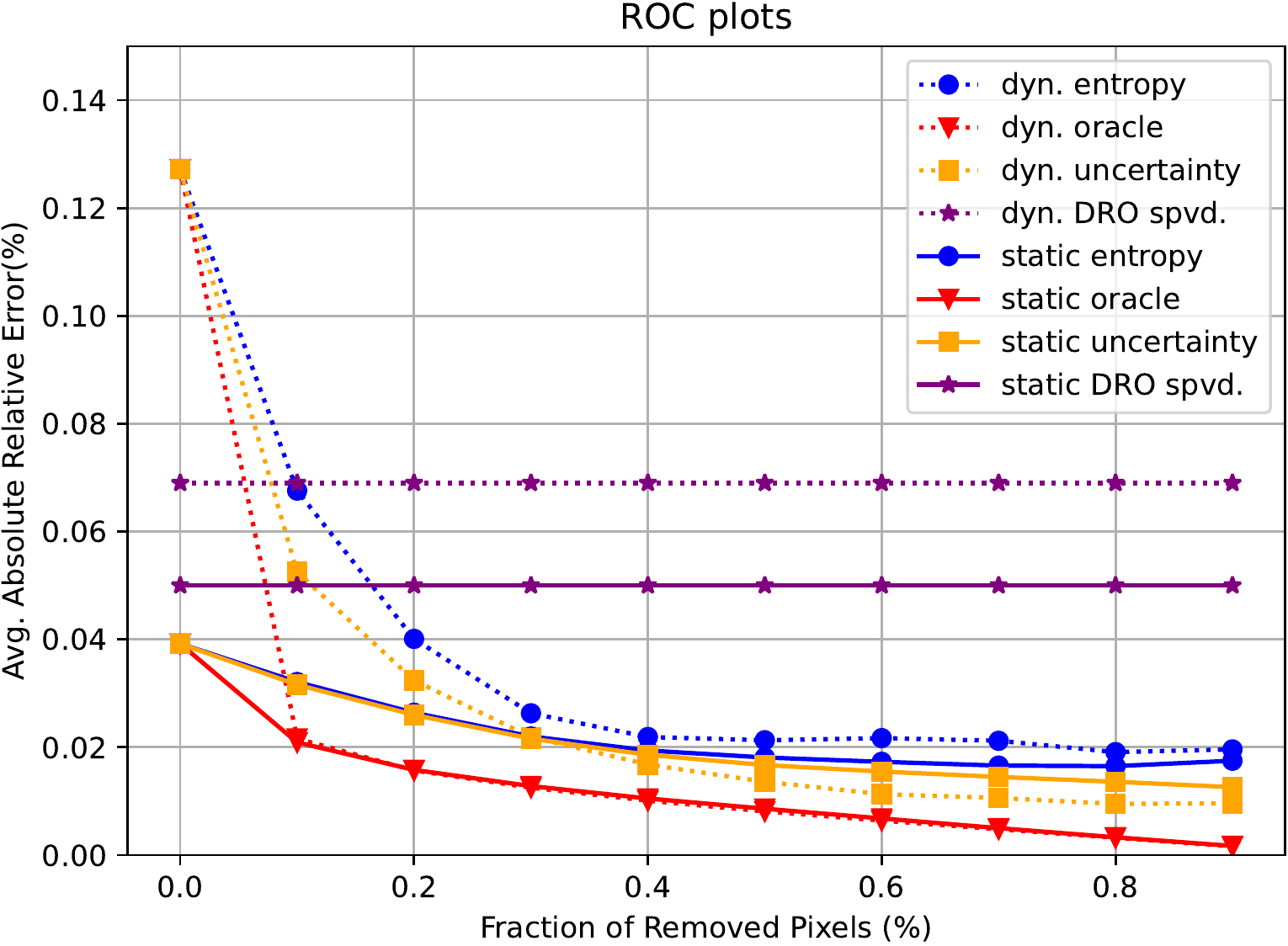}
		\caption{DeepV2cD* on KITTI 2015 train split}
	\end{subfigure}
	\caption{
		Sparsification plots on KITTI 2012 and 2015: We sort the pixels in ascending order of uncertainty. The x-axis represents the percentage of pixels, that are included in the evaluation and the y-axis represents the absolute relative error (ARE). An oracle acts as upper bound for identifying errors. We also include a random selection baseline for the Eigen split. We can distinguish between errors on moving objects and static background on KITTI 2015. We include supervised DRO \cite{gu2021dro} for comparison. 
		While the Shannon entropy is already useful for filtering, the learned uncertainty correlates better with the actual error. By rejecting few percent of the pixels, we can reliably filter out gross outliers and achieve SotA performance. The filtered predictions of DeepV2cD converge to similar reconstruction performance for both static and dynamic parts of the scene.
	}
	\label{fig:kitti_roc}
\end{figure*}

\subsection{Uncertainty}
The previous experiments have shown, that with an uncertainty head and an improved regularization strategy DeepV2cD achieves SotA depth prediction accuracy on 5-frame videos on KITTI. In Figure \ref{fig:uncertainty}, we show qualitative examples of our learned uncertainty on the KITTI dataset. It correlates with hard to match pixels, occlusions, reflective surfaces and areas of missing ground truth. In general it is higher at far away pixels in the scene. Since the uncertainty head can be run in parallel to the depth prediction head, this does not come with an additional computational overhead.

In the next experiments, we investigate the quality of the learned uncertainty for identifying errors made by the model. This implicates, that sorting the pixels by uncertainty results in the same ordering as when we sort by errors. We use sparsification plots \cite{hu2012quantitative,zhang2020adaptive,yang2019inferring} and compare the learned uncertainty with three baselines: 1. Shannon entropy 2. Random filtering 3. Oracle. Furthermore, we compare DeepV2cD to the inherent uncertainty of DeepV2D. Figure \ref{fig:kitti_roc} shows, that DeepV2cD can learn a reliable aleatoric uncertainty while achieving SotA accuracy. We beat all baselines and can achieve cleaner reconstructions after filtering with no significant overhead. When throwing away $ 20\% $ of the pixels, the avg. abs. rel. error is below $ 2.5 \% $. Related work DRO \cite{gu2021dro} achieves SotA accuracy by collapsing the cost volume, but cannot detect the model errors inherent to the sparse supervision. We argue that this property is very useful for downstream tasks, such as map building or scene flow estimation \cite{brickwedde2019mono}.

\section{Conclusion}
\label{sec:conclusion}
In this paper, we investigated the performance of deep SfM frameworks on several autonomous driving datasets. We show improved results for a cost volume based architecture due to better loss supervision and an additional uncertainty head. Our results indicate, that supervised models make errors on real datasets mainly due to a lack of supervision and that they are able to learn more accurate depths when provided with high quality dense data on virtual datasets. Missing supervision can be compensated by considering the aleatoric uncertainty. Instead of taking the Shannon entropy inside the cost volume, a learned uncertainty showed better performance at identifying outliers. 

\section{ACKNOWLEDGEMENT}
Research presented here has been supported by the Robert Bosch GmbH. We thank our colleauges Jan Fabian Schmidt, Annika Hagemann and Holger Janssen for fruitful discussions and proof reading.

%
%
\bibliographystyle{splncs04}
\bibliography{draft}

\end{document}